\pdfoutput=1
\documentclass[11pt]{article}

\usepackage[final]{acl}
\usepackage{times}
\usepackage{latexsym}

\usepackage[T1]{fontenc}
\usepackage[utf8]{inputenc}

\usepackage{microtype}

\usepackage{inconsolata}

\usepackage{graphicx}
\usepackage{amsmath}
\usepackage{subcaption}
\usepackage{booktabs}
\usepackage{array}
\usepackage{amsfonts}
\usepackage{multirow}
\usepackage{makecell}
\usepackage{amssymb}
\usepackage{xcolor}
\usepackage{xspace}
\usepackage{longtable}
\usepackage{tabularx}
\usepackage{enumitem}

\definecolor{stage1}{HTML}{F091A0} %E89BAA, D15485
\definecolor{stage2}{HTML}{DB7729}
\definecolor{stage3}{HTML}{55B5C2}
\definecolor{stage4}{HTML}{F4BC41}

\definecolor{a1}{HTML}{E2D3BF}
\definecolor{r1}{HTML}{BEB8DA}
\definecolor{p1}{HTML}{EB8675}
\definecolor{a2}{HTML}{BAA188}
\definecolor{r2}{HTML}{9588C3}
\definecolor{p2}{HTML}{E3503C}

\usepackage{listings}
\usepackage{xcolor}
\newcommand{\emojimentalhealth}{\includegraphics[height=1.4em, trim=0 0.7em 0 0]{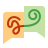}}

\newcommand{\esconv}{{ESConv}$^{\circ}$\xspace}
\newcommand{\ntconv}{{NTConv}$^{\circ}$\xspace}
\title{\emojimentalhealth Reframe Your Life Story: Interactive Narrative Therapist and Innovative Moment Assessment with Large Language Models}

\author{
 \textbf{Yi Feng\textsuperscript{1,2}}\thanks{Work done during internship at the CoAI Group.}
 \textbf{Jiaqi Wang\textsuperscript{3}}
 \textbf{Wenxuan Zhang\textsuperscript{4}}
 \textbf{Zhuang Chen\textsuperscript{5}}
 \textbf{Shen Yutong\textsuperscript{6}}\\
 \textbf{Xiyao Xiao\textsuperscript{7}}
 \textbf{Minlie Huang\textsuperscript{2,7}}\footnotemark[2]
 \textbf{Liping Jing\textsuperscript{1}}\thanks{Corresponding author}
 \textbf{Jian Yu\textsuperscript{1}}\\
 \textsuperscript{1}Beijing Key Laboratory of Traffic Data Mining and Embodied Intelligence,\\ Beijing Jiaotong University
 \textsuperscript{2}CoAI Group, DCST, IAI, BNRIST, Tsinghua University\\
 \textsuperscript{3}Tencent Jarvis Lab
 \textsuperscript{4}Singapore University of Technology and Design\\
 \textsuperscript{5}Central South University
 \textsuperscript{6}Tsinghua University
 \textsuperscript{7}Lingxin AI\\
 yifeng@bjtu.edu.cn 
}

\begin{document}
\maketitle

\begin{abstract}
Recent progress in large language models (LLMs) has opened new possibilities for mental health support, yet current approaches lack realism in simulating specialized psychotherapy and fail to capture therapeutic progression over time. Narrative therapy, which helps individuals transform problematic life stories into empowering alternatives, remains underutilized due to limited access and social stigma.
We address these limitations through a comprehensive framework with two core components. First, \textit{INT} (Interactive Narrative Therapist) simulates expert narrative therapists by planning therapeutic stages, guiding reflection levels, and generating contextually appropriate expert-like responses. Second, \textit{IMA} (Innovative Moment Assessment) provides a therapy-centric evaluation method that quantifies effectiveness by tracking ``Innovative Moments'' (IMs), critical narrative shifts in client speech signaling therapy progress. 
Experimental results on 260 simulated clients and 230 human participants 
reveal that \textit{INT} consistently outperforms standard LLMs in therapeutic quality and depth.
We further demonstrate the effectiveness of \textit{INT} in synthesizing high-quality support conversations to facilitate social applications\footnote{The Code and data will be released at \url{https://github.com/MIMIFY/narrative-therapy-llm}.}.

\end{abstract}

\section{Introduction}
\begin{figure}[t]
  \includegraphics[width=0.96\columnwidth]{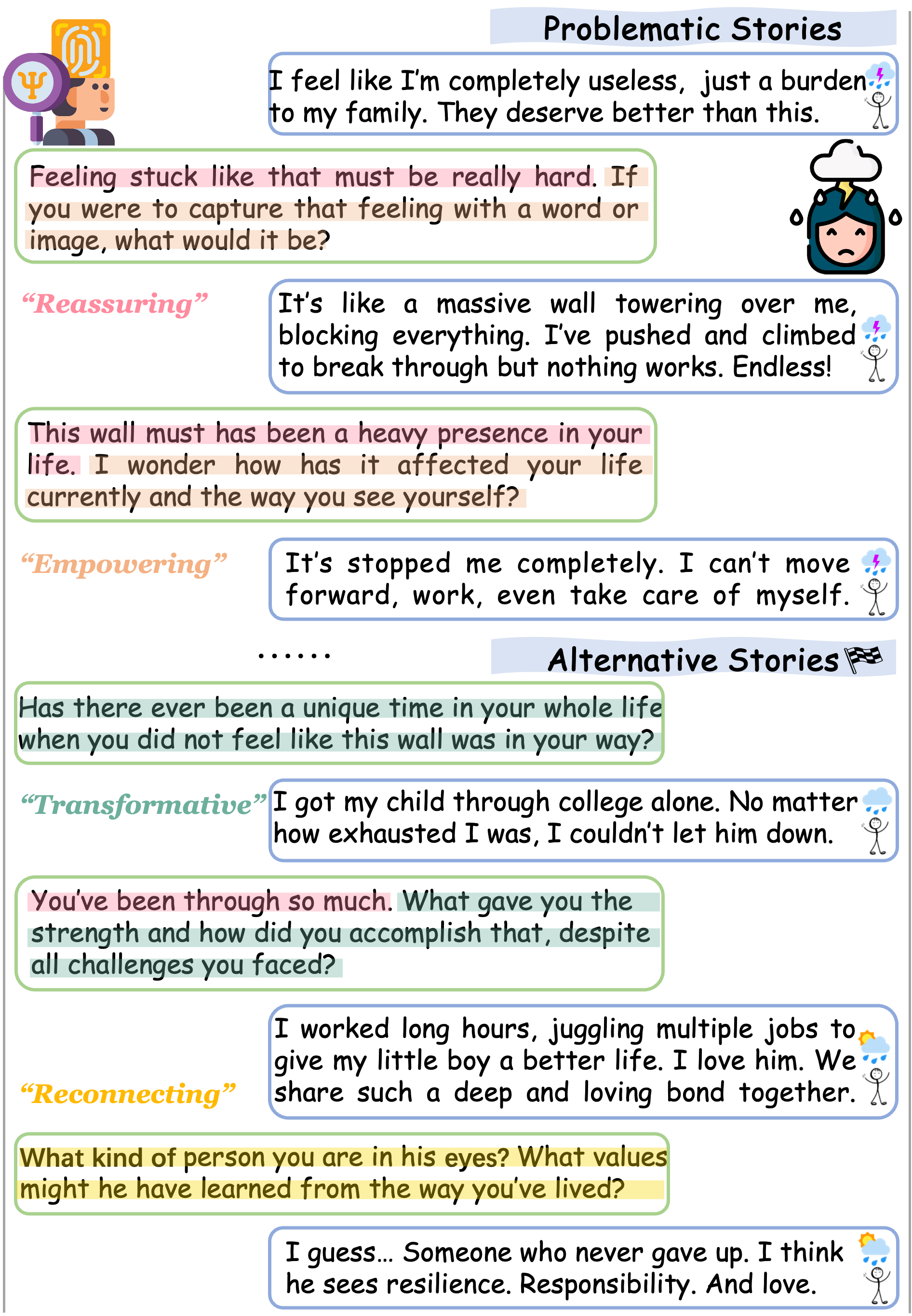}
  \caption{An example dialogue shows how narrative therapy can help a single mother shift from viewing her disability and life as a despairing burden (\textit{problematic}) to rediscovering purpose by recognizing her resilience in overcoming adversity
  %to support her son and build a nurturing bond 
  (\textit{alternative}).}
  \label{fig:introduction}
\end{figure}

Human experience is fundamentally organized via life stories. Narrative cognition, how we shape and interpret these personal stories, thus plays a crucial role across psychological domains including memory, emotion, cognition, and social functioning \citep{sarbin1986narrative, angus2004handbook}, and serves as a key perspective for understanding and treating mental health issues, as it enables individuals to construct self-identity through these stories \citep{bruner1985chapter, fletcher2023storythinking}.

However, psychological distress often distorts these stories into problematic narratives, reinforcing negative identities and vicious cycles \citep{white1990narrative}. When individuals become trapped in these problem-saturated narratives, they struggle to access alternative perspectives and more adaptive self-conceptions.
Therefore, evidence-based narrative therapy helps reframe such distorted self-stories through a structured process, including \textit{reassuring} clients, \textit{empowering} them to externalize problems, \textit{reconnecting} them with relationships, and facilitating \textit{transformative} ``Innovative Moments'' (IMs)--signs of therapeutic progress where clients construct alternative narratives that challenge old patterns of passivity, shame, and self-devaluation \citep{white1990narrative, gonccalves2012innovative}. 
Figure \ref{fig:introduction} illustrates this therapeutic transformation process, showing how problematic narratives could be reshaped through structured therapeutic intervention.

Recent advances in large language models (LLMs) offer promising solutions for assisting a wide range of emotional support and general therapeutic conversation tasks \citep{chen-etal-2023-soulchat, xie2024psydt, qiu2024interactive}. However, most existing approaches rely on broad counseling heuristics or surface-level imitation, without alignment to structured therapeutic frameworks like empirically grounded narrative therapy. Simply applying those methods often leads to unrealistic interactions, where compliant simulated clients and formulaic responses replace authentic exploration \citep{louie-etal-2024-roleplay, carik2025reimagining}. Furthermore, current evaluation methods for therapeutic dialogues depend on reference-based metrics or generic dialogue qualities (e.g., ``empathy''), fail to capture the longitudinal progression of therapeutic change.

In this work, we propose a novel LLM-based approach for simulating and evaluating professional narrative therapy. Specifically, we introduce the \textit{Interactive Narrative Therapist} (\textit{\textbf{INT}}) for simulation and the \textit{Innovative Moment Assessment} (\textit{\textbf{IMA}}) for tracking and evaluating therapeutic progression, both grounded in core psychotherapy principles. 

(1) \textit{INT} is a planning-based workflow that simulates professional narrative therapists in a theory-driven, progression-aware manner. To the best of our knowledge, we are the first to translate narrative therapy principles into a computational framework by systematically formalizing the therapeutic process into four progressive stages (i.e., from \textit{reassuring} to \textit{reconnecting}, Figure \ref{fig:stage-reflection}) and structured reflection levels within each stage. \textit{INT} drives the session forward by explicitly planning the therapeutic state (stage and reflection level), then generating responses aligned with expert style through retrieval-augmented generation.

(2) \textit{IMA} provides a  novel therapy-centric evaluation approach that operationalizes ``Innovative Moment'' (IM) theory for computational assessment. \textit{IMA} identifies six IM types across two levels (Table \ref{tab:imcs}) to capture narrative transformation in client speech. We further quantify therapeutic effectiveness via proposed IM salience metric, the proportion of IM related client speech to the entire dialogue. \textit{IMA} enables, for the first time, longitudinal tracking of narrative shifts throughout therapy sessions, offering a process-oriented assessment aligned with clinical outcomes.

We conduct comprehensive evaluations via both automated assessment (with 260 simulated clients) and extensive human evaluation (involving 230 participants in the main (200) and follow-up (30) studies, plus expert annotators). Results consistently demonstrate that our \textit{INT} outperforms all standard LLMs based on direct role-playing, with significant gains across core therapeutic dimensions and nearly doubling the elicitation of advanced narrative transformation markers according to \textit{IMA}. 
We further demonstrate the effectiveness of \textit{INT} in synthesizing support conversations of high quality to facilitate realistic social applications.

\begin{figure*}[t]
  \includegraphics[width=\textwidth]{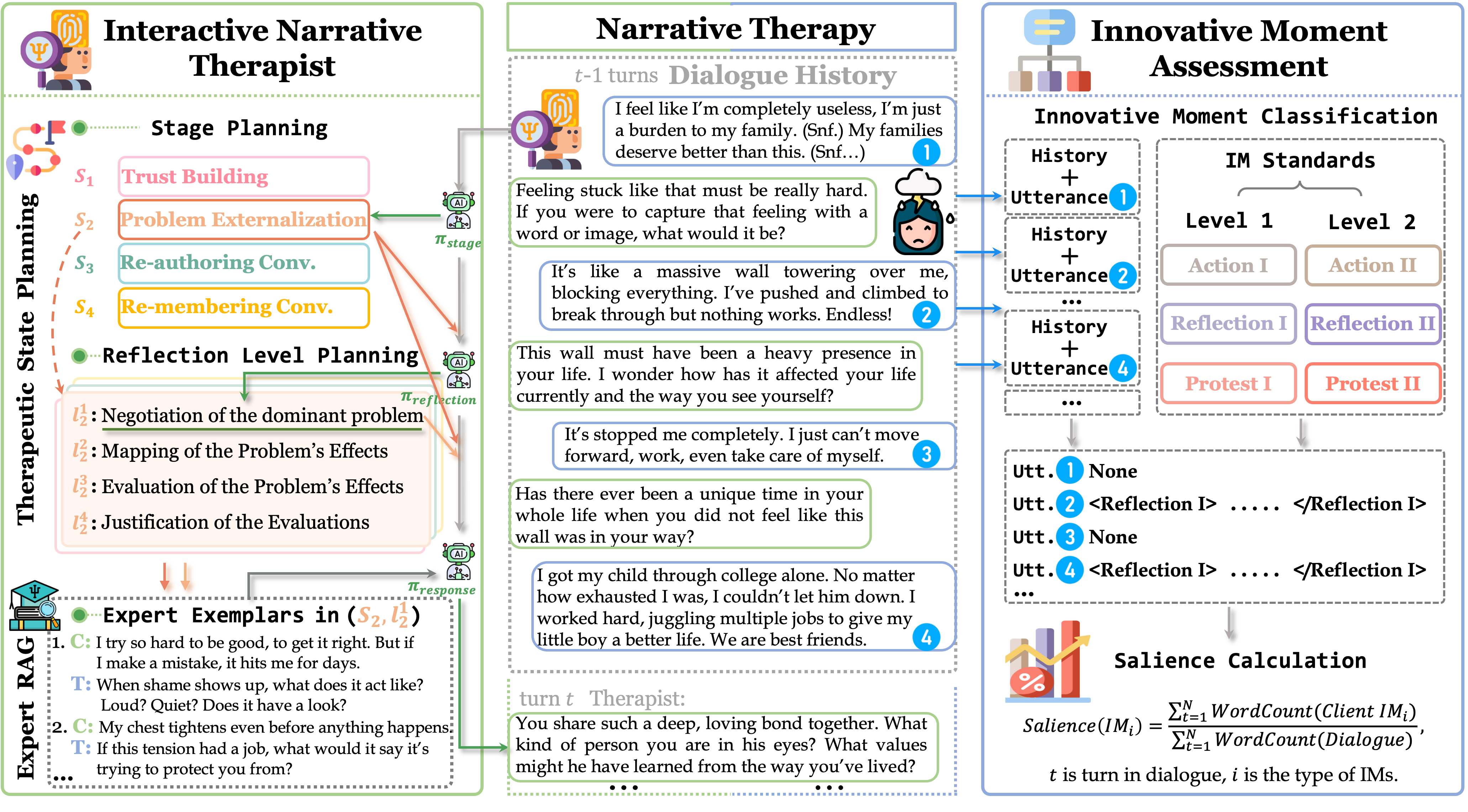}
  \caption{Overview of our framework, comprising \textit{INT} (left, green box) for narrative therapist simulation and \textit{IMA} (right, blue box) for evaluating therapeutic progression. \textit{INT} plans the therapeutic state (stage, reflection level) and generates responses aligned with expert style using retrieved exemplars. \textit{IMA} classifies each client utterance into six IM types (e.g., Action I) across two levels and quantifies therapeutic progression via salience calculation.}
  \label{fig:method}
\end{figure*}

\section{Related Work}
\noindent \textbf{LLMs for Mental Health Support}
Recent work has explored LLMs for psychotherapy simulation, ranging from empathetic dialogue datasets (e.g., ESConv \citealp{liu-etal-2021-towards}, PsyQA \citealp{sun2021psyqa}) to synthetic role-played conversations \citep{chenpersona, qiu2024interactive, zhang2024cpsycoun}. While scalable, such methods often lack therapeutic fidelity, producing overly compliant clients and generic therapists \citep{louie-etal-2024-roleplay, carik2025reimagining}. Most rely on surface-level imitation without modeling therapeutic structure. In contrast, our \textit{INT} specifically targets narrative therapy with dynamic state planning to enable theory-driven, progression-aware therapy simulation that more closely approximates authentic therapeutic practice.

\noindent \textbf{Evaluation for Therapeutic Dialogue}
Conventional metrics like BLEU \citep{papineni2002bleu}, ROUGE \citep{lin2004rouge} and BERTScore \citep{zhangbertscore} fail to capture therapeutic progression. Recent approaches (e.g., empathy detection \citep{sharma-etal-2020-computational}, qualitative indicators rating \citep{jin2023psyeval}, LLM-based Working Alliance Inventory scoring \citep{qiu2024interactive}) assess overall quality but overlook in-session progression. These methods typically focus on static therapeutic dimensions rather than the dynamic evolution of client narratives. We address this via human evaluation on grounded therapeutic dimensions \citep{white2007maps} and expert annotation of Innovative Moments \citep{gonccalves2012innovative}, tracking client transformation across entire dialogues.

\section{Methodology}
\label{sec:methodology}

This section outlines our framework (Figure \ref{fig:method}), including the general notation (\S\ref{subsec:problem_formulation}), the planning-based narrative therapist simulation approach \textit{INT} (\S\ref{subsec:INT}), and the therapy-centric evaluation method \textit{IMA} grounded in Innovative Moments (\S\ref{subsec:IMC}).

\subsection{Problem and Notation}
\label{subsec:problem_formulation}
Let $\mathcal{D} = {(C_i, T_i)}_{i=1}^N$ represent a therapeutic conversation session, where $C_i$ and $T_i$ denote the client utterance and therapist's response at turn $i$.
Given the dialogue history $\mathcal{H}_t = {(C_i, T_i)}_{i=1}^{t-1}$ and the current client utterance $C_t$, the simulated therapist aims to generate an appropriate response $T_t$ that adheres to narrative therapy principles (Figure\ref{fig:stage-reflection}). For evaluation, the goal is to classify the Innovative Moments (Table \ref{tab:imcs}) of client utterances $C_{i=1}^N$ given the entire session $\mathcal{D}$.

\subsection{INT for Simulation}
\label{subsec:INT}
\subsubsection{Theoretical Principles}
\label{subsubsec:theory_NT}

\begin{figure*}[t]
  \includegraphics[width=\textwidth]{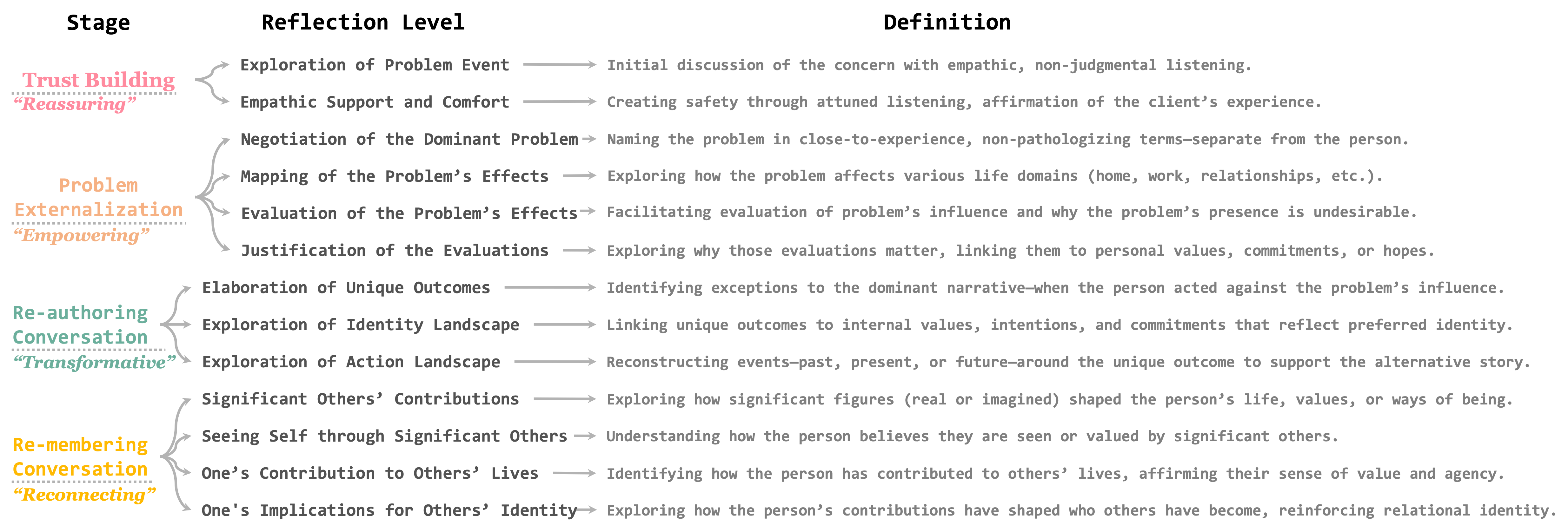}
  \caption{Overview of reflection levels within each therapeutic stage of narrative therapy, including specific definitions and examples of therapist utterances.}
  \label{fig:stage-reflection}
\end{figure*}

Maintaining an appropriate therapeutic pace is critical: premature progression may ruin client trust, whereas stagnation impedes therapeutic progress \citep{madigan2011narrative}. Expert narrative therapists strategically guide conversations through different therapeutic stages while dynamically adjusting their reflection depth of guidance based on client readiness \citep{white2007maps}. 
To navigate this balance between client readiness and therapeutic progression, we pioneer a structured simulation of professional narrative therapists. Specifically, we translate narrative therapy principles \citep{white1990narrative, white2007maps} into a computational framework \textit{INT} by systematically formalizing the therapeutic process into four progressive stages and associated reflection levels, as illustrated in Figure \ref{fig:stage-reflection}.

We define a formal planning space $\Phi = (\mathcal{S}, \mathcal{L})$ that captures the hierarchical organization of narrative therapy, where: 

$\mathcal{S} = \{s_1, s_2, s_3, s_4\}$ denotes therapeutic stages: Trust Building ($s_1$, Reassuring), Problem Externalization ($s_2$, Empowering), Re-authoring Conversation ($s_3$, Transformative), and Re-membering Conversation ($s_4$, Reconnecting). 

$\mathcal{L}_i=\{l_i^1, l_i^2,...,l_i^{n_i}\}$ denotes $n_i$ reflection levels within each stage $S_i$, representing increasing depths of therapeutic engagement.

\subsubsection{Therapeutic State Planning}

\label{subsubsec:state_plan_NT}
\noindent\textbf{Stage Planning:} For each dialogue turn $t$, we first determine the appropriate therapeutic stage $s_t \in \mathcal{S}$ based on the dialogue history $\mathcal{H}_t = {(C_i, T_i)}_{i=1}^{t-1}$ and current client utterance $C_t$. We implement a planning function $\Psi_S(\cdot)$ using a stage-tracking instruction $\pi_{stage}$ with an LLM:
\begin{equation}
s_t = \Psi_S(\mathcal{H}_t, C_t, \pi_{stage}).
\end{equation}
The stage-tracking instruction identifies client trust levels, problem internalization signs, emerging alternative narratives, and relationship references to determine the appropriate therapeutic stage.

\noindent\textbf{Reflection Level Planning:} Once the therapeutic stage $s_t$ is determined, we select the appropriate reflection level $l_t^t \in \mathcal{L}_t$ within that stage:
\begin{equation}
l_t^t = \Psi_L(s_t, \mathcal{H}_t, C_t, \pi_{reflection}).
\end{equation}
The reflection-tracking $\pi_{reflection}$ guides the LLM to determine the appropriate depth of therapeutic engagement based on client narrative position, emotional readiness, and previous reflection levels.

\begin{table*}[t]
\centering
\small
\setlength{\tabcolsep}{4pt}
\renewcommand{\arraystretch}{1.2}
\resizebox{\textwidth}{!}{

\begin{tabular}{p{1.6cm}|p{16.4cm}}
\hline
\textbf{IM Type} & \multicolumn{1}{c}{\textbf{Definition  \&  Contents}} \\ \hline

\multicolumn{2}{c}{\textbf{Level 1: Creating Distance from the Problem}} \\ \hline
\textcolor{a1}{Action I} & 
\textbf{Contents:} New behavioral strategies to overcome the problem, active exploration of solutions and information about the problem. \newline
\underline{\textit{Problem stories:}} Client got nervous and refused to go to public places after experiencing domestic violence from partner.\newline
\underline{\textit{Client utterance:}} \textcolor{a1}{<Action I>}Yesterday I went out to the cinema for the first time this month to watch a movie.\textcolor{a1}{</Action I>} \\ 
% \hline

\textcolor{r1}{Reflection I} & 
\textbf{Contents:} New understandings of the problem, intention to fight (CONTEST) the problem’s demands, and references to self-worth. \newline
\underline{\textit{Problem stories:}} Client let the depression take over his/her life for a long time.\newline
\underline{\textit{Client utterance:}} \textcolor{r1}{<Reflection I>}It wants to control my entire life, eventually taking it all away.\textcolor{r1}{</Reflection I>} \\ 
% \hline

\textcolor{p1}{Protest I} & 
\textbf{Contents:} Rejecting or objecting to the problem, critique of those who support it, and critique of problematic facets of the self. \newline
\underline{\textit{Problem stories:}} Client must live according to his/her parents' expectations.\newline
\underline{\textit{Client utterance:}} \textcolor{p1}{<Protest I>}Parents should love their children, not constantly judge them. I've really had enough.\textcolor{p1}{</Protest I>} \\ \hline

\multicolumn{2}{c}{\textbf{Level 2: Centered on the Change}} \\ \hline

\textcolor{a2}{Action II}  & \textbf{Contents:} Generalization into the future and other life dimensions of good outcomes (performed or projected actions). \newline
\underline{\textit{Problem stories:}} Client was afraid to say no even he/she was uncomfortable.\newline
\underline{\textit{Client utterance:}} \textcolor{a2}{<Action II>}I'll also bring this boundary awareness to work, like no longer working overtime silently.\textcolor{a2}{</Action II>} \\ 
% \hline

\textcolor{r2}{Reflection II} & \textbf{Contents:} Contrasting self (what changed?) or self-transformation (how/why change occurred?). \newline
\underline{\textit{Problem stories:}} Client exhibited excessive anxiety when coping with daily pressure.\newline
\underline{\textit{Client utterance:}} \textcolor{r2}{<Reflection II>}Before, when I encountered any problem, I would spend the whole day anxious, self-critical, even wanting to escape. ...I suddenly realized I'm not so easily defeated anymore.\textcolor{r2}{</Reflection II>}\\ 
% \hline

\textcolor{p2}{Protest II} & 
\textbf{Contents:} Centering on the self, affirming personal rights, needs, and values. \newline
\underline{\textit{Problem stories:}}  Client exhibited the pattern of deriving self-worth from prioritizing others' needs.\newline
\underline{\textit{Client utterance:}} \textcolor{p2}{<Protest II>}I think my feelings are important too. I have the right to say 'no', the right to rest when tired, rather than constantly pleasing others. I want to start living for myself, not according to others' expectations.\textcolor{p2}{</Protest II>} \\ \hline

\end{tabular}
}
\caption{Classification of six Innovative Moments (IMs) across Level 1 and Level 2 in psychotherapy.}
\label{tab:imcs}
\end{table*}

\subsubsection{Retrieval-Augmented Responding} 
\label{subsubsec:rag_NT}
After determining the therapeutic state ($s_t$, $l_t^t$), we retrieve top-$k$ relevant expert exemplar responses $\mathcal{E}_t = \{e_1, \ldots, e_k\}$ ($k=5$ in this paper) from a predefined repository $\mathcal{E}$, using cosine similarity\footnote{We use OpenAI’s embedding api: \url{https://platform.openai.com/docs/api-reference/embeddings} to obtain the dense vector representations.} between $e_m$ and state-augmented query $(C_t, s_t, l_t^t)$. We use $\mathcal{E}_t$ to augment the response generation:
\begin{equation}
T_t = \Psi_T(\mathcal{H}_t, C_t, s_t, l_t^t, \mathcal{E}_t; \pi_{response}),
\end{equation}
where $\pi_{response}$ instructs the LLM to generate a response aligned with the determined stage $s_t$, reflection level $l_t^t$, and exemplar style of expert $\mathcal{E}_t$.

\subsection{IMA for Evaluation}
\label{subsec:IMC}

\subsubsection{Theoretical Principles}
\label{subsubsec:theory_MA}
 
Innovative Moments (IMs) are episodes where clients express thoughts, feelings, or behaviors that contradict their problem-saturated narratives \citep{white2007maps}. These moments indicate therapeutic progress as clients develop alternative narratives that challenge problematic self-perceptions. 
Psychotherapy research consistently shows that clients who progress from early-stage IMs (e.g., resistance or awareness) to advanced IMs involving personal change and self-redefinition tend to achieve more successful therapeutic outcomes \citep{gonccalves2012innovative, gonccalves2024innovative}. Therefore we operationalize such theory for computational assessment and propose \textit{IMA} inspired by \citet{gonccalves2011tracking} and \citet{montesano2017self} to evaluate therapeutic effectiveness explicitly by specifically tracking narrative shifts of Innovative Moments.
Table \ref{tab:imcs} (full version in Appendix \ref{appendix:expert_annotation}) outlines the six summarized IM types with distinct markers that signal shifts from problem-saturated narratives to preferred alternatives, along with examples illustrating the annotation process. We formalize these IMs here as $\mathcal{I} = \{\mathcal{IM}_1, \mathcal{IM}_2\}$, where:

$\mathcal{IM}_1 = \{A_1, R_1, P_1\}$ represents Level 1 IMs (Creating Distance from the Problem), corresponding to the early stages of Trust Building and Problem Externalization: Action I (new behaviors), Reflection I (new understandings), and Protest I (objections to the problem).

$\mathcal{IM}_2 = \{A_2, R_2, P_2\}$ represents Level 2 IMs (Centered on the Change) - the advanced stages of Re-authoring and Re-membering: Action II (future-oriented behaviors), Reflection II (self-transformation), and Protest II (empowerment).

\subsubsection{Innovative Moment Classification}
\label{subsubsec:imcs_MA}
For each client utterance $C_t$, we identify the presence and type of Innovative Moments using a multi-label classification approach:
\begin{equation}
\mathcal{I}_t = \Gamma(C_t, \mathcal{H}_t; \pi_{IM}),
\end{equation}
where $\mathcal{I}_t \subseteq \{\mathcal{IM}_1,\mathcal{IM}_2\}$ represents the set of IMs detected in the client utterance, and $\pi_{IM}$ is a detection instruction that guides the LLM or human experts to analyze the narrative for specific transformation markers as detailed in Table \ref{tab:imcs}.

\subsubsection{Therapeutic Effectiveness Measurement}
\label{subsubsec:salience_MA}
To quantify therapeutic effectiveness, we use a salience metric that measures the proportion of client's speech associated with a particular IM type, which is recommended in psychotherapy research.

\begin{equation}
\text{Salience}(\mathcal{I}_i) = \frac{\sum_{t=1}^{N} \text{WordCount}(C_t \cap \mathcal{I}_i)}{\sum_{t=1}^{N} \text{WordCount}(C_t \cup T_t)},
\label{eq:salience}
\end{equation}
where $\text{WordCount}(C_t \cap \mathcal{I}_i)$ represents the number of words in client utterance $C_t$ that are classified as containing IM type $\mathcal{I}_i$, and $\text{WordCount}(C_t \cup T_t)$ represents total words of the $t_{th}$ turn conversation.

This metric allows us to compare the relative prevalence of different IM types, track progression from Level 1 to Level 2 IMs, and evaluate the overall effectiveness of different therapeutic approaches in eliciting narrative transformation.

\begin{table*}[t]
\centering
\small
\setlength{\tabcolsep}{3pt}
\renewcommand{\arraystretch}{1.2}
\resizebox{\textwidth}{!}{
\begin{tabular}{l|cccc|c|ccc|ccc|c}
\hline
\multicolumn{1}{c|}{\multirow{2}{*}{\textbf{Model}}} & \multicolumn{5}{c|}{\textbf{Therapeutic Dimensions}} & \multicolumn{7}{c}{\textbf{Innovative Moment Assessment(Salience)}} \\
\cline{2-13}
& \textbf{Reas.} & \textbf{Emp.} & \textbf{Trans.} & \textbf{Recon.} & \textbf{Avg.}
& \textbf{Action I} & \textbf{Reflection I} & \textbf{Protest I}
& \textbf{Action II} & \textbf{Reflection II} & \textbf{Protest II} & \textbf{SUM} \\
% \hline
% \multicolumn{13}{c}{\textbf{Role-playing}} \\
\hline

Claude-3.7-sonnet & 3.13 & 3.29 & 3.12 & 2.96  & 3.13 & 2.459\% & 6.796\% & 0.036\% & 4.762\%	& 8.971\%& 0.100\% & 23.124\%  \\
Gemini-2.5-pro & 2.18 & 2.47 & 2.84 & 2.63 & 2.53 & 3.982\% & 7.656\% & 0.027\%	& 8.782\%	& 15.738\% & 0.117\%	&36.302\%  \\
Qwen-2.5 & 3.51 & 3.35 & 3.08 & 3.10  & 3.26 & 3.740\% & 7.460\% & 0.011\% & 7.328\% & 12.819\% & 0.051\%	& 31.409\%  \\
GLM-4-plus & 2.93 & 3.58 & 3.23 & 3.17   & 3.23 & 4.602\% & 8.933\% & 0.062\%	& 8.169\%	& 15.504\% & \textbf{0.148\%}	& \textbf{37.418\%}  \\
Deepseek-V3 & 3.31 & 3.80 & 3.71 & 3.45  & 3.57 & 3.824\% &  \textbf{9.388\%} & 0.092\% & 8.099\%	& 14.760\% & 0.067\% & 36.234\%  \\
Doubao-1.5-pro & 2.80 & 3.23 & 3.00 & 2.95 & 3.00 & \textbf{4.866\%}	& 8.489\%	 & 0.082\%	& 10.606\% & 17.988\% & 0.079\%	&42.110\%  \\
GPT-4o  & 3.34 & 3.52 & 3.19 & 3.19  & 3.31 & 3.115\%	& 7.480\% & 0.037\% & 6.819\% & 11.770\% & 0.127\% & 29.348\%  \\
\hline
\textit{INT} & \textbf{3.60} & \textbf{3.87} & \textbf{3.84} & \textbf{3.51} & \textbf{3.71} &1.594\% & 3.092\% & \textbf{0.096\%} & \textbf{11.136\%} & \textbf{19.072\%} & 0.074\% & 35.064\% \\
\hline

\end{tabular}
}
\caption{Model-only assessment with GPT-4o simulated clients across therapeutic dimensions and IM salience.
}
\label{tab:dialogue-eval-full-gpt}
\end{table*}
\begin{table*}[t]
\centering
\small
\setlength{\tabcolsep}{3pt}
\renewcommand{\arraystretch}{1.2}
\resizebox{\textwidth}{!}{
\begin{tabular}{l|cccc|c|ccc|ccc|c}
\hline
\multicolumn{1}{c|}{\multirow{2}{*}{\textbf{Model}}} & \multicolumn{5}{c|}{\textbf{Therapeutic Dimensions}} & \multicolumn{7}{c}{\textbf{Innovative Moment Assessment(Salience)}} \\
\cline{2-13}
& \textbf{Reas.} & \textbf{Emp.} & \textbf{Trans.} & \textbf{Recon.} & \textbf{Avg.}
& \textbf{Action I} & \textbf{Reflection I} & \textbf{Protest I}
& \textbf{Action II} & \textbf{Reflection II} & \textbf{Protest II} & \textbf{SUM} \\
\hline
\multicolumn{13}{c}{\textbf{Role-playing}} \\
\hline
Claude-3.7-sonnet & 3.08 & 2.77 & 2.56 & 2.40  & 2.70 & 3.539\% & 6.895\% & 0.629\% & 4.059\% & 5.919\% & 0.794\% & 21.835\% \\
Gemini-2.5-pro & 3.01 & 2.14 & 2.01 & 1.94 &   2.28 & 3.458\% & 6.934\% & 0.486\% & 3.281\% & 4.647\% & 0.613\% & 19.419\% \\
Qwen-2.5 & 2.76 & 2.37 & 2.15 & 2.10 &  2.35 & 3.171\% & 6.355\% & 0.538\% & 3.551\% & 4.971\% & 0.679\% & 19.265\% \\
GLM-4-plus & 2.83 & 2.70 & 2.27 & 2.27 &  2.52 & 3.251\% & 6.606\% & 0.613\% & 3.839\% & 5.248\% & 0.774\% & 20.331\% \\
Deepseek-V3 & 2.73 & 2.54 & 2.46 & 2.61  & 2.59 & 3.136\% & 6.722\% & 0.577\% & 3.737\% & 5.225\% & 0.728\% & 20.125\% \\
Doubao-1.5-pro & 2.66 & 2.45 & 2.10 & 2.10 &  2.33 & 3.056\% & 6.528\% & 0.556\% & 3.551\% & 4.855\% & 0.702\% & 19.248\% \\
GPT-4o  & 3.11 & 2.75 & 2.52 & 2.49 & 2.72 & 3.513\% & 6.895\% & 0.624\% & 4.211\% & 5.827\% & 0.788\% & 21.858\% \\
\hline
\multicolumn{13}{c}{\textbf{Ours}} \\
\hline
\textit{INT} & 3.09 & \textbf{3.11} & \textbf{3.42} & \textbf{3.37} & \textbf{3.25} & 2.794\% & 6.834\% & \textbf{0.662\%} & \textbf{8.730\%} & \textbf{9.680\%} & \textbf{0.998\%} & \textbf{29.698\%} \\
\textit{w/o RAG} & 3.13 & 2.92 & 2.74 & 2.69 &  2.87 & \textbf{3.573\%} & \textbf{8.333\%} & 0.610\% & 4.235\% & 9.438\% & 0.803\% & 26.992\% \\
\textit{w/o RAGRL} & \textbf{3.16} & 2.83 & 2.65 & 2.63  & 2.82 & 3.135\% & 5.010\% & 0.309\% & 6.488\% & 6.934\% & 0.586\% & 22.462\% \\
\hline
\end{tabular}
}
\caption{Human interactive evaluation with participants, annotators across therapeutic dimensions and IM salience.
}
\label{tab:dialogue-eval-full}
\end{table*}

\section{Experiments}
\label{sec:experiments}
\begin{figure*}[t]
  \includegraphics[width=0.49\linewidth]{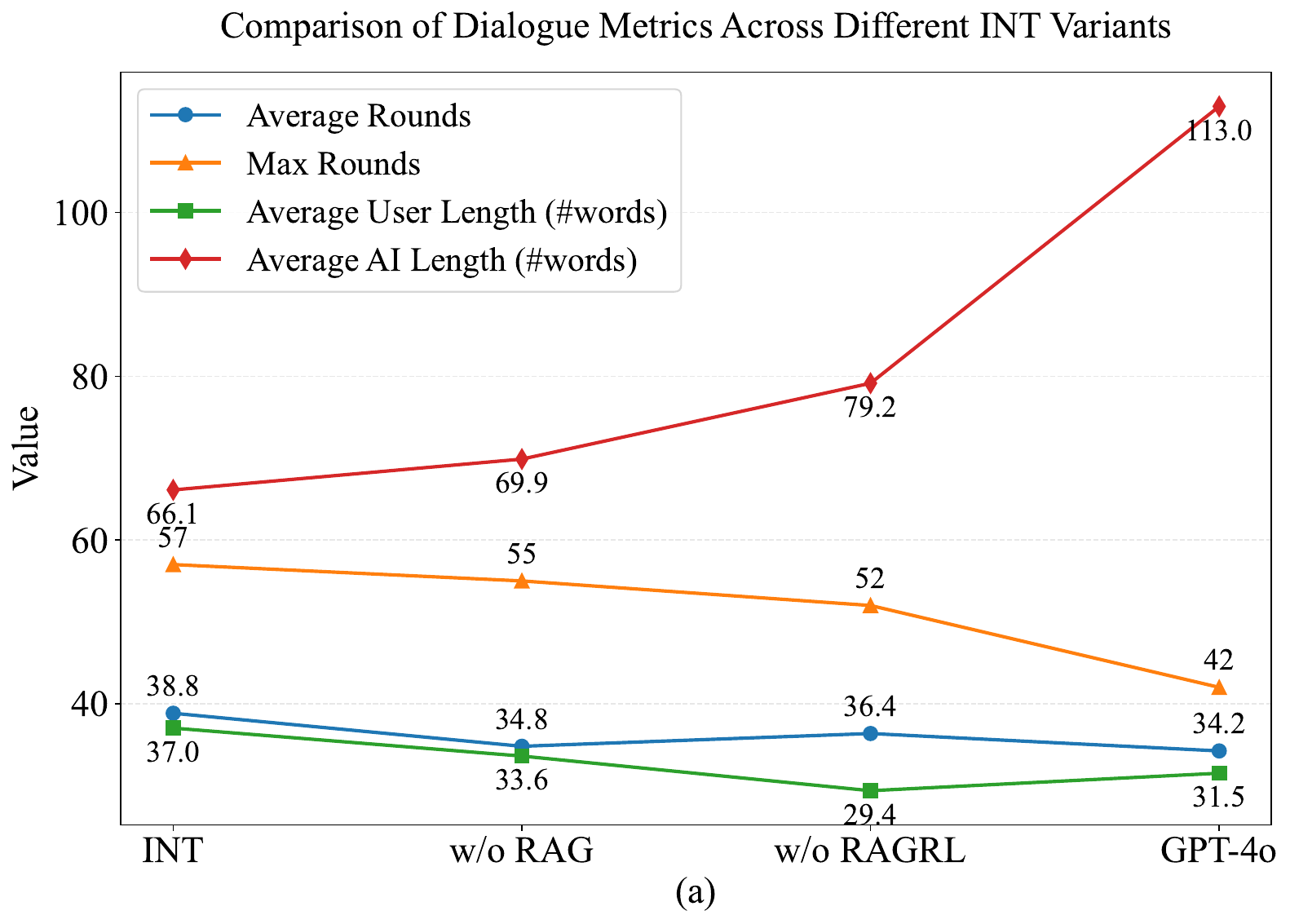} \hfill
  \includegraphics[width=0.49\linewidth]{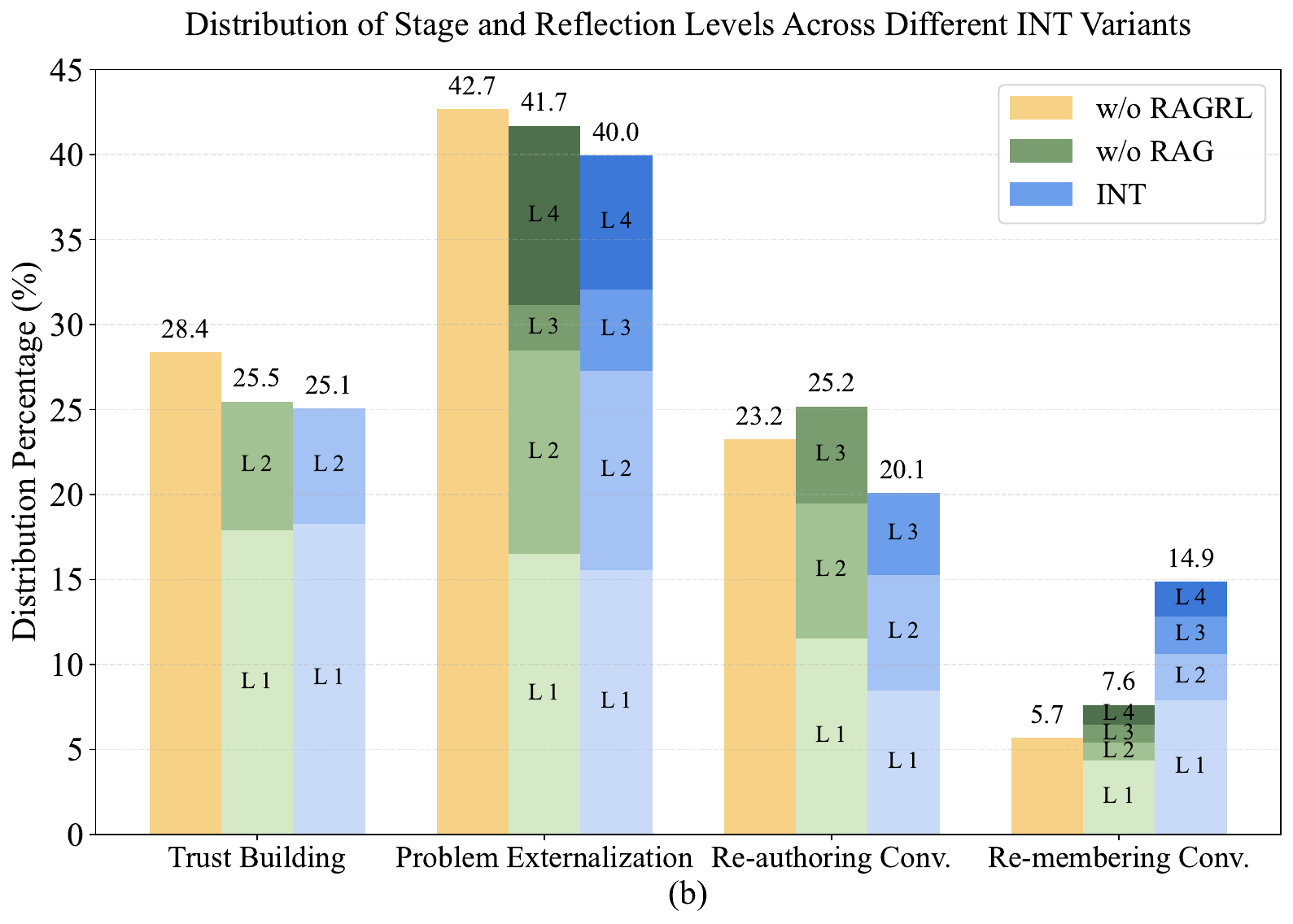}
  \caption{Statistics (a) and therapeutic state distribution (b) of interactive dialogues with INT variants. In (b), yellow/green/blue series show variants with color gradients indicating reflection levels within each therapeutic stage.}
  \label{fig:ablation}
\end{figure*}

\subsection{Experimental Setup}
\subsubsection{Variants and Models}
For comparative evaluation, we examine a range of advanced LLMs: GPT-4o \citep{hurst2024gpt}, Claude-3.7-sonnet \citep{anthropic2024}, Gemini-2.5-pro \citep{gemini2024}, Qwen-2.5-72B-instruct (``Qwen-2.5'') \citep{yang2024qwen2}, GLM-4-plus \citep{glm2024chatglm}, DeepSeek-V3 \citep{liu2024deepseek}, and Doubao-1.5-pro-32k (``Doubao-1.5-pro'') \citep{doubao2024}. 
All receive identical instructions carefully structured by core therapy principles and detailed response guidelines (Appendix \ref{sec:INT-instructions}).

Preliminary studies show GPT-4o's superior performance in following complex therapeutic setting and reasoning capabilities and we select it to implement our \textit{INT} framework.\footnote{We accessed GPT-4o via: \url{https://openai.com/api/}. See Appendix \ref{appendix:implementation} for query parameters, Appendix \ref{sec:INT-instructions} for $\pi_{stage},\pi_{reflection},\pi_{response}$, and role-playing instructions.} To study the component contributions, we design three variants: 
(1) \textit{INT} incorporates complete therapeutic state planning (stage and reflection level) plus retrieval-augmented generation; 
(2) \textit{INT w/o RAG} maintains planning but removes retrieval augmentation; 
(3) \textit{INT w/o RAGRL} implements only stage planning without reflection level planning.

\subsubsection{Evaluation Metrics} 
\label{subsubsec:metrics}

\noindent\textbf{Therapeutic Dimension Assessment:} We evaluate systems across 4 therapeutic dimensions using a 5-point Likert scale:
\textit{Reassuring} (creating safety for client disclosure, \texttt{Reas.}),
\textit{Empowering} (facilitating problem externalization, \texttt{Emp.}),
\textit{Transformative} (uncovering alternative narratives, \texttt{Trans.}),
\textit{Reconnecting} (strengthening significant relationships, \texttt{Recon.}).The \textit{Average} score is (\texttt{Avg.}).

\noindent \textbf{Innovative Moment Assessment:} Following the established protocols from \citet{gonccalves2011tracking}, each client utterance is annotated with a subset of six IM categories (Table \ref{tab:imcs}), or labeled as ``None'' if no IM is present.
For co-occurrence cases, we follow explicit coding rules in psychotherapy: when ``Action'' and ``Reflection'' markers co-occur, both are coded; when either co-occurs with ``Protest'', the utterance is coded as ``Protest''. For each multi-turn dialogue session, we finally compute the salience of IMs using Equation \eqref{eq:salience}, then average these scores across all sessions within the same model to report system-level results.

\subsubsection{Interaction Setting}
\label{subsubsec:client}
To assess how \textit{INT} compares to direct role-playing approaches, we conducted both model-only and human interactions and evaluations.

\noindent\textbf{Interaction with Simulated Clients:} 
To enable systematic comparison under identical conditions, we conduct automatic evaluation with simulated clients\footnote{See Appendix \ref{appendix:model-only-evaluation} for detailed instructions of client simulation and GPT evaluation.}: 
(1) We use GPT-4o to simulate clients based on 260 real-world client profiles (20\%) extracted from the authentic ESConv dataset \citep{liu-etal-2021-towards}, preserving each seeker's demographic information, background story, emotional state, and core concerns.
(2) Each simulated client then engages in conversations for at least 35 turns with each system.
(3) In the end, we use GPT-4o to simulate a strict counseling supervisor to evaluate the resulting dialogues across all therapeutic dimensions and identified IMs in each client utterance following the theoretical protocol.
This fully automated approach controls client variability, and ensures rigorous and consistent evaluation through a simulated expert supervisor.

\noindent\textbf{Interaction with Human Participants:} 
To assess realistic effectiveness, we conduct a human evaluation with 200 participants (20 per system, gender-balanced, aged 18--30, mostly college students) who provide informed consent and are assured of privacy protection and anonymity. Participants are randomly assigned to one blind system and engage in a therapeutic conversation about personal challenges with it for at least 30 minutes ($\geq$ 30 turns on average). After that, they are asked to rate their experience on four dimensions using a 5-point Likert scale. After completion, participants are fully debriefed about the research objectives.\footnote{See Appendix \ref{appendix:human_eval} for complete experimental protocol, participant demographics, and questionnaire details.}

For \textit{IMA}, two qualitative researchers (master-level with backgrounds in narrative therapy) are trained by \citet{gonccalves2011tracking} protocol until achieving high inter-rater reliability (Cohen's $\kappa > 0.75$). 
They independently annotate 10 randomly sampled sessions per system (15 minutes/session on average), using explicit rules (\S\ref{subsubsec:metrics}) with all client identifiers removed. Disagreements are resolved through discussion, and unresolved cases ($\sim3\%$) are adjudicated by the senior therapist.\footnote{See Appendix \ref{appendix:expert_annotation} for coder training and annotating details.}

\subsection{Effectiveness Over Role-Playing}
Tables \ref{tab:dialogue-eval-full-gpt} and \ref{tab:dialogue-eval-full} present the results from automated GPT evaluation and human assessment, offering comprehensive insights into our \textit{INT}'s performance compared to direct role-playing models.

\noindent \textbf{Consistent Excellence:} 
\textit{INT} outperforms all role-playing models consistently across core narrative therapy dimensions. In human evaluation, it achieves superior scores in Emp. (3.11 vs. next best 2.77), Trans. (3.42 vs. 2.56), and Recon. (3.37 vs. 2.61), dimensions directly aligned with narrative therapy's core processes: problem externalization, re-authoring, and re-membering. 
Simulated client evaluations show a similar trend (3.87, 3.84, 3.51 in these dimensions), confirming effectiveness across settings.
For \textit{IMA} salience, both evaluation approaches confirm \textit{INT}'s superiority in facilitating narrative transformation. \textit{INT} achieves the highest overall IM salience in human evaluation (29.698\% vs. next best 21.858\%), and excels at eliciting advanced Level 2 IMs with highest salience in $A_2, R_2$ categories on both evaluation approaches.  

\noindent \textbf{Reliability Across Interaction Settings:} 
As shown in Table \ref{tab:dialogue-eval-full-gpt}, Deepseek-V3 performs well with simulated clients but degrades with real clients, suggesting they may excel in more predictable, compliant simulation environments but struggle with the complexity and variability of real human interactions. 
Besides, GPT-based evaluation also tends to overestimate dimension scores and advanced IMs, underscoring the importance of expert annotation for accurate therapeutic assessment.
Table \ref{tab:dialogue-eval-full-gpt} reveals that models like Doubao-1.5-pro and Gemini-2.5-pro score high on IM salience but low on therapeutic dimensions. This gap likely stems from simulated clients being too compliant, leading to surface-level progress that lacks real therapeutic depth.
In contrast, \textit{INT} maintains strong performance across both settings, indicating robust adaptability to diverse interaction styles. 

\noindent \textbf{Beyond Comfort to Transformation:}
Human evaluation provides more nuanced insights. Claude-3.7-sonnet and GPT-4o achieve high Reassuring scores (3.08, 3.11) but significantly underperform in later therapeutic stages, particularly Transformative (2.56, 2.52) and Reconnecting (2.40, 2.49). This suggests these models prioritize emotional support over narrative reshaping skills—a common pitfall in mental health applications.
Interestingly, ablated versions of our framework (\textit{w/o RAG} and \textit{w/o RAGRL}) still outperform most baseline models, especially in later therapeutic stages. This progressive improvement from \textit{w/o RAGRL} to \textit{w/o RAG} to \textit{INT} (Trans.: 2.65→2.74→3.42; Recon.: 2.63→2.69→3.37) underscores the cumulative value of each component in our framework.

\subsection{Human-Centered Analysis}  %In-Depth Human Interaction Analysis
Given the observed discrepancies between GPT and human evaluations, we focus our subsequent analysis on human interactions as they provide more realistic and challenging test conditions.

\noindent\textbf{Component Contribution Analysis:} 
We conduct a within-subject study with 30 participants (aged 19–42, M = 27.4; 50\% female), following the same ethical procedures (\S\ref{subsubsec:client}). Each participant interacts with all \textit{INT} variants on the same personal topic, allowing comparison under controlled conditions.
Figure~\ref{fig:ablation}(a) shows \textit{INT} produces focused responses (66.1 vs. 113.0 words), longer dialogues (57 vs. 42 turns), and richer user input (38.8 vs. 31.5 words) than directly role-playing with GPT-4o. As shown in Figure~\ref{fig:ablation}(b), while all variants focus on Problem Externalization (40\%), \textit{INT} advances further into Re-authoring (20.1\%) and Re-membering (14.9\%), with more frequent deep reflection levels (L3/L4), especially in later stages.
Table~\ref{tab:dialogue-eval-full} confirms this trend: Emp., Trans., and Recon. scores steadily rise from role-playing agent to \textit{INT} (+0.36, +0.90, +0.88), highlighting the impact of each component on therapeutic effectiveness.

\noindent \textbf{Therapeutic Progression Analysis:}
\begin{figure}[t]
  \includegraphics[width=\columnwidth]{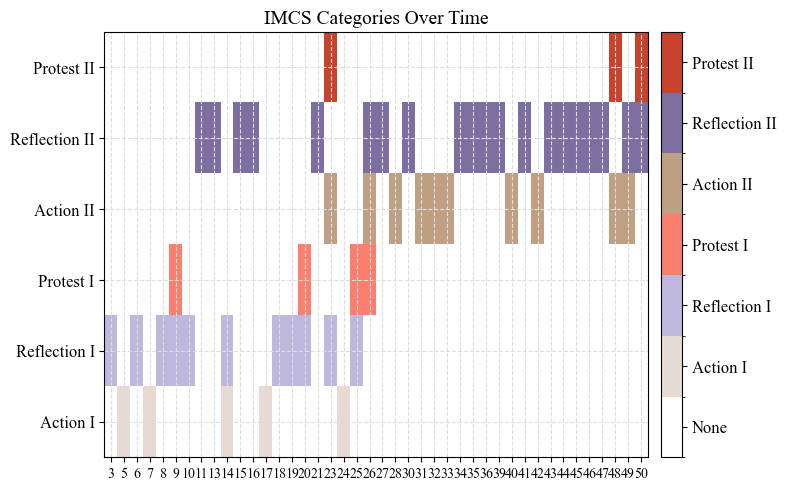}
  \caption{The IMs trajectory across dialogue turns (x-axis) in a randomly selected dialogue session from \textit{INT}.}
  \label{fig:imcs-case-study}
\end{figure}
Figure~\ref{fig:imcs-case-study} visualizes the trajectory of narrative transformation across dialogue turns in a representative \textit{INT} session, demonstrating a clear three-phase progression that mirrors established therapeutic change patterns:
Early Phase (Turns 3–20): Dominated by Level 1 IMs, especially Reflection I, as the client begins to reconsider problem narratives.
Middle Phase (Turns 21–35): Rise of Level 2 IMs (Action II, Reflection II) signaling a shift toward alternative narratives and concrete changes.
Late Phase (Turns 36–50): Sustained Level 2 IMs with occasional Protest II, indicating growing empowerment and narrative reconstruction as clients develop stronger agency and self-advocacy.
This Level $1\to2$ shift aligns with patterns in successful therapy \citep{montesano2017self}, supporting effectiveness of \textit{INT}. 
Notably, the frequent co-occurrence of Action II and Reflection II in consecutive turns suggests that as clients re-imagine themselves, 
they simultaneously discover new possibilities for action.

\begin{table}[t]
\centering
\small
\setlength{\tabcolsep}{3pt}
\renewcommand{\arraystretch}{1.2}
\resizebox{0.9\columnwidth}{!}{
\begin{tabular}{l|cc|cc}
\hline

\multicolumn{1}{l|}{\textbf{Metric}} & \multicolumn{2}{c|}{\textbf{ESConv-test}} & \multicolumn{2}{c}{\textbf{NTConv-test}} \\
\cline{1-5}
\multicolumn{1}{l|}{\textbf{Automatic}} & \textbf{\esconv} & \textbf{\ntconv} & \textbf{\esconv} & \textbf{\ntconv}  \\
\hline
\textbf{BLEU-1} $\uparrow$ & 16.38 & 17.08 & 23.35 & 34.48  \\
\textbf{BLEU-2} $\uparrow$ & 6.77 & 6.89 & 8.15 & 14.86  \\
\textbf{ROUGE-L} $\uparrow$ & 14.56 & 15.65 & 17.95 & 25.44  \\
\textbf{METEOR} $\uparrow$ & 15.42 & 18.56 & 18.54 & 29.04  \\
\textbf{Extrema} $\uparrow$ & 49.62 & 49.94 & 47.86 & 50.73  \\
\hline
\multicolumn{1}{l|}{\textbf{Human}} & \multicolumn{4}{c}{\textbf{\ntconv  vs. \esconv (Win: Loss: Tie)}} \\
\hline
\multicolumn{1}{l|}{\textbf{Fluency}} & \multicolumn{2}{c|}{19 : 15 : 16} & \multicolumn{2}{c}{28 : 10 : 12} \\
\multicolumn{1}{l|}{\textbf{Identification}} & \multicolumn{2}{c|}{24 : 8 : 18} & \multicolumn{2}{c}{31 : 8 : 11} \\
\multicolumn{1}{l|}{\textbf{Comforting}} & \multicolumn{2}{c|}{ 15: 13 : 22} & \multicolumn{2}{c}{26 : 12 : 12} \\
\multicolumn{1}{l|}{\textbf{Suggestion}} & \multicolumn{2}{c|}{11 : 14 :25} & \multicolumn{2}{c}{35 : 6 : 9} \\
\multicolumn{1}{l|}{\textbf{Overall}} & \multicolumn{2}{c|}{20 : 16 : 14} & \multicolumn{2}{c}{32 : 8 : 10} \\
\hline
\end{tabular}
}
\caption{Automatic and Human evaluation results of \esconv and \ntconv on both test sets.}
\label{tab:esconv}
\end{table}

\subsection{Empowering Social Application}
\label{subsec:downstream}
We further explore whether exposure to narrative therapy principles can improve models' effectiveness in real-world help-seeking scenarios, specifically asking whether incorporating narrative therapy structure enhances LLMs’ performance in emotional support tasks.
For each ESConv dialogue, we create a corresponding NTConv dialogue by having simulated clients (following the same protocol as \S\ref{subsubsec:client}) interact with our \textit{INT} framework. The resulting NTConv dataset maintains one-to-one correspondence with ESConv while incorporating systematic narrative therapy principles, and both datasets are split 8:2 into train/test sets.

Qwen3-8B \citep{Qwen2025} is fine-tuned on each dataset under identical settings\footnote{See Appendix \ref{appendix:implementation} for training parameters}. The resulting models \esconv and \ntconv are evaluated on both test sets using standard metrics: BLEU-1/2, ROUGE-L, METEOR \citep{banerjee2005meteor}, and Extrema \citep{liu2016not}. Human assessment by three experts on 50 sampled test dialogues determines the preferred model via majority vote.
Table \ref{tab:esconv} shows \ntconv outperforms \esconv across all metrics on both test sets. On ESConv data, \ntconv achieves higher BLEU-1 (17.08 vs. 16.38) and METEOR (18.56 vs. 15.42), while \esconv substantially underperforms on NTConv test data. Human evaluation also confirms that \textit{INT} can facilitate high-quality support conversations in realistic help-seeking scenarios, demonstrating its potential for broader social applications beyond clinical therapeutic settings.

\section{Conclusion}
We propose the first comprehensive framework translating narrative therapy principles into AI practice through \textit{INT} and \textit{IMA}. 
The \textit{INT} systematically formalizes therapeutic progression through explicit stage planning and reflection level guidance, enabling more authentic therapeutic interactions. The \textit{IMA} is a therapy-centric evaluation approach to specifically quantify therapeutic effectiveness through tracking narrative shifts in client speech. 
Comprehensive experiments with 260 simulated clients and 230 human participants demonstrate that \textit{INT} significantly outperforms standard role-playing methods in therapeutic quality and depth.
We further demonstrate the effectiveness of \textit{INT} in synthesizing high-quality support conversations for broader social applications.

\section*{Limitations}
In this work, we introduce \textit{INT} and \textit{IMA}, an innovative theory-grounded framework for simulating professional narrative therapist and evaluating effective therapeutic progression with LLMs. Although our experimental results demonstrate the viability and effectiveness of our approach, several limitations need to be considered.

\noindent\textbf{Cross-Cultural Applicability:}
Our framework is developed and evaluated primarily within English-speaking contexts, potentially limiting its cross-cultural applicability. Narrative structures and therapeutic norms vary across cultures, requiring adaptations for global deployment.

\noindent \textbf{Therapeutic Complexity:}
In real-life scenarios, psychological counseling is highly complex. Therapeutic effectiveness depends not only on empathetic responses but also on the appropriate timing and application of specialized techniques. If techniques like problem externalization or narrative reconstruction are applied at inappropriate moments in the therapeutic journey, they may impede rather than facilitate progress. Our current model, while structured, may not fully capture this delicate balance between different therapeutic strategies.

\noindent \textbf{Technical Constraints:}
The current implementation relies on GPT-4o, which may pose scalability challenges for resource-constrained applications. Additionally, while our approach is more clinically grounded than previous methods, further longitudinal studies are needed to confirm whether narrative transformation markers translate to measurable well-being outcomes. Future work should address these limitations through multilingual adaptation, more nuanced timing mechanisms for therapeutic techniques, more efficient model implementations, and extended clinical validation.

\section*{Ethics Statement}
\noindent\textbf{Participant Protection:}
Our research prioritizes participant welfare and data privacy. All participants were fully informed about the study purpose, data collection procedures, and confidentiality measures before providing written informed consent. Participants were compensated fairly and debriefed thoroughly after completion. We implemented comprehensive safeguards including anonymization of all identifiable information and secure data storage protocols.
The research team include a senior therapist to ensure clinical appropriateness and prevent potential harm.

We conducted rigorous data sanitization procedures to safeguard privacy. Following established data copyright practices, we make our evaluation dataset available only for research purposes.

\noindent\textbf{Mental Health Applications Considerations}
We acknowledge that AI systems for mental health support carry inherent risks. It is critical to emphasize that our virtual dialogue agent cannot and should not replace professional therapeutic care. We designed our system to complement rather than substitute human therapists, particularly for complex clinical cases requiring specialized judgment.
Users must be clearly informed that they are interacting with an AI system, and responses should be used only as references.

By sharing our methodology and findings, we aim to advance responsible innovation in mental health technology while maintaining the highest ethical standards in participant protection, data handling, and system design.

\section*{Acknowledgments}
This work was partly supported by the National Key Research and Development Program of China under Grant 2024YFE0202900; the NSFC projects 62441614; the National Natural Science Foundation of China under Grant (62436001, 62176020); the Joint Foundation of the Ministry of Education for Innovation team (8091B042235); the Fundamental Research Funds for the Central Universities (2019JBZ110); and the State Key Laboratory of Rail Traffic Control and Safety (Contract No.RCS2023K006), Beijing Jiaotong University. 
We sincerely thank our collaborators for their invaluable contributions, including insightful feedback and support in refining the paper and addressing key challenges. We thank all participants and experts in this paper for their patience and enthusiasm. We thank the anonymous reviewers for carefully reading our paper and their insightful comments and suggestions.

\bibliography{acl25_narrative_therapy}

\appendix

\section{Implementation Details}
\label{appendix:implementation}
To ensure the reproducibility of our experiments, we provide the following essential details regarding the model, dataset, software, and training configurations used in \S\ref{subsec:downstream} (Table \ref{tab:reproducibility}).

Table \ref{tab:sft-inference} presents the hyper-parameters employed for training and testing NTConv, ESConv in the supervised fine-tuning (SFT) and the inference stage (\S\ref{subsec:downstream}). These hyper-parameters include the learning rate, the LORA settings, the inference settings that are crucial for replicating our results and understanding the performance differences we have observed between the models.

We use different sets of hyper-parameters when querying the GPT-4 API for various purposes. These hyper-parameters are found to work well with the GPT-4 model (``GPT-4o'' engine). We listed them in Table \ref{tab:query-api}. The current rates are \$0.03 per 1000 tokens for input (prompt) and \$0.06 per 1000 tokens for output (completion). Based on these prices and the scope of our experiments, the total cost for interaction and evaluation experiment is approximately \$500. 

\begin{table*}[htbp]
\small
\begin{tabularx}{\linewidth}{@{}lX@{}}
\toprule
\textbf{Item} & \textbf{Description} \\
\midrule
Model and Data & We use Qwen3-8B for the downstream application with dataset ESConv and NTConv. \\

\addlinespace
Framework Versions & The transformers library version is 4.51.3, deepspeed is 0.16.7, datasets is 3.5.0 and jieba is 0.42.1. \\
\addlinespace
Model Parameters & LoRA fine-tuning was applied to the Qwen3-8B model. During inference, the model with LoRA adapters typically requires around 14GB of GPU memory when using 16-bit precision. \\
\addlinespace
Training Time & \ntconv \& \esconv are trained with global steps of 30,000 and a torch\_dtype of ``float16". The total training time is approximately 12 hours. \\
\addlinespace
Package Versions & python=3.11, torch=2.7.0, cuda=12.6. \\
\bottomrule
\end{tabularx}
\caption{Reproducibility Checklist}
\label{tab:reproducibility}
\end{table*}

\begin{table*}[tbp]
  \centering
  \caption{The hyper-parameters we use respectively for Supervised Fine-tuning (SFT) and Inference.}
  \resizebox{0.95\textwidth}{!}{
    \begin{tabular}{c|cccccc}
    \toprule
    \multicolumn{1}{c|}{\multirow{2}[4]{*}{SFT}} & num\_train\_epochs & learning\_rate & lora\_alpha & lora\_dropout & lora\_rank & target\_modules \\
\cmidrule{2-7}          & 20     & 5$e^{-5}$ & 16    & 0     & 8     & v\_proj, q\_proj \\
    \midrule
    \multirow{2}[4]{*}{Inference} & max\_new\_tokens & top\_p & temperature & num\_beams & repetition\_penalty & length\_penalty \\
\cmidrule{2-7}          & 1024  & 0.9   & 0.7  & 1     & 1.0     & 1.0 \\
    \bottomrule
    \end{tabular}%
    }
  \label{tab:sft-inference}%
\end{table*}%

\begin{table*}[tbp]
\centering
\scriptsize
\renewcommand{\arraystretch}{0.95}
\caption{Hyper-parameters for querying OpenAI API in different experiments.}
\resizebox{0.98\textwidth}{!}{
\begin{tabular}{p{2.3cm}p{0.6cm}p{0.6cm}p{0.65cm}p{0.65cm}p{0.65cm}p{0.65cm}c}
\midrule
\centering  Experiments &  \centering  Temp. & \centering Top\_P & Freq. Penalty &  Pres. Penalty  & Beam Size  &  Max Length &   Stop Sequences \\
\midrule
Therapist Role-playing & 1.0 & 0.95 & 0 & 0 & 1 & 300 & None (default) \\
Client Simulation & 0.7 & 0.5 & 0 & 2.0 & 1 & 300 & None (default) \\
Stage Planning & 0.5 & 1.0 & 0 & 2.0 & 1 & 200 & None (default) \\
Reflection Planning & 0.5 & 1.0 & 0 & 2.0 & 1 & 200 & None (default) \\
Response Generation & 0.8 & 0.9 & 0 & 1.5 & 1 & 300 & None (default) \\
IM\_Annotation & 0.1 & 1.0 & 0.2 & 0 & 1 & 512 & None (default) \\
Dimension Evaluation & 0.1 & 1.0 & 0.2 & 0 & 1 & 512 & None (default) \\
\midrule
\end{tabular}%
}
\label{tab:query-api}%
\end{table*}%

\section{Interactive Narrative Therapist \textit{INT}}
\label{sec:INT-instructions}
We construct a stage–reflection–level mapping table along with the representative expert responses in the Table \ref{tab:stage-reflection}.

\begin{table*}[t]
\centering
\small
\setlength{\tabcolsep}{4pt}
\renewcommand{\arraystretch}{1.2}
\resizebox{\textwidth}{!}{
\begin{tabular}{|>{\centering\arraybackslash}p{3cm}|p{7.2cm}|p{7.2cm}|}
% \begin{tabular}{|p{3.1cm}|p{7.2cm}|p{7.2cm}|}
\hline
\textbf{Reflection Level} $\mathcal{L}$ & \multicolumn{1}{c}{\textbf{Reflection Definition}} & \multicolumn{1}{|c|}{\textbf{Example} $\mathcal{E}$} \\ \hline

\multicolumn{3}{|c|}{\textcolor{stage1}{$\mathcal{S}_1$ \textbf{Stage I: Trust Building}} \texttt{----> \textbf{Reassuring}}} \\ \hline
\textbf{L1}: Exploration of Problem Event  & Initial discussion of the concern with empathic, non-judgmental listening. & “Can you tell me what brought you here today?” \\ \hline
\textbf{L2}: Empathic Support and Comfort & Creating safety through attuned listening, affirmation, and validation of the client’s experience. & “That sounds really tough. I appreciate you sharing that with me.” \\ \hline

\multicolumn{3}{|c|}{\textcolor{stage2}{$\mathcal{S}_2$ \textbf{Stage II: Problem Externalization}} \texttt{----> \textbf{Empowering}}} \\ \hline
\textbf{L1}: Negotiation of the Dominant Problem  & Naming the problem in close-to-experience, non-pathologizing terms—separate from the person. & “So it sounds like ‘Self-Doubt’ has been interfering with your decisions lately—does that sound right?” \\ \hline
\textbf{L2}: Mapping of the Problem’s Effects & Exploring how the problem affects various life domains (home, work, relationships, etc.). & “When anxiety shows up, how does it affect your workday or how you relate to friends?” \\ \hline
\textbf{L3}: Evaluation of the Problem’s Effects & Facilitating evaluation of the problem’s influence and why the problem’s presence is undesirable. & “What do you think about the way Anger has been affecting your relationship with your partner?” \\ \hline
\textbf{L4}: Justification of the Evaluations & Exploring why those evaluations matter, linking them to personal values, commitments, or hopes. & “You said this matters because you value connection and mutual respect—can you tell me more about that?” \\ \hline

\multicolumn{3}{|c|}{\textcolor{stage3}{$\mathcal{S}_3$} \textcolor{stage3}{\textbf{Stage III: Re-authoring Conversation}} \texttt{----> \textbf{Transformative}}} \\ \hline
\textbf{L1}: Elaboration of Unique Outcomes & Identifying exceptions to the dominant narrative—when the person acted against the problem’s influence. & “You stood up to fear that day—what made that possible?” \\ \hline
\textbf{L2}: Exploration of Identity Landscape & Linking unique outcomes to internal values, intentions, and commitments that reflect preferred identity. & “When you held firm to your views about changing your phone number, what were you hoping for?” \\ \hline
\textbf{L3}: Exploration of Action Landscape & Reconstructing events—past, present, or future—around the unique outcome to support the alternative story. & “What might I have seen you doing then that showed these hopes mattered to you?” \\ \hline

\multicolumn{3}{|c|}{\textcolor{stage4}{$\mathcal{S}_4$} \textbf{\textcolor{stage4}{Stage IV: Re-membering Conversation}} \texttt{----> \textbf{Reconnecting}}} \\ \hline
\textbf{L1}: Significant Others’ Contributions & Exploring how significant figures (real or imagined) shaped the person’s life, values, or ways of being. & “How did your grandmother influence the values you hold today?” \\ \hline
\textbf{L2}: Seeing Self through Significant Others & Understanding how the person believes they are seen or valued by significant others. & “How do you think she would describe who you are and what you stand for?” \\ \hline
\textbf{L3}: One’s Contribution to Others’ Lives & Identifying how the person has contributed meaningfully to others’ lives, affirming their sense of value and agency. & “What do you think your support meant to your younger sibling during that time?” \\ \hline
\textbf{L4}: One's Implications for Others’ Identity & Exploring how the person’s contributions have shaped who others have become, reinforcing their relational identity. & “What did staying with her then make possible—for how she sees herself, as someone who can persevere, maybe?” \\ \hline

\end{tabular}
}
\caption{Overview of reflection levels within each therapeutic stage of narrative therapy, including specific definitions and examples of therapist utterances.}
\label{tab:stage-reflection}
\end{table*}

\subsection{Stage Planning Instructions}
The stage-tracking prompt $\pi_{stage}$ (Table \ref{tab:stage-prompt}) guides the LLM to analyze the dialogue history and client utterance in terms of narrative therapy principles, as outlined in Table \ref{tab:stage-reflection}. It incorporates explicit rules for stage transitions based on client narrative indicators and output the determined stage.

\begin{table*}[htbp]
\small
\renewcommand{\arraystretch}{0.95}
\begin{tabular}{p{15cm}}
\toprule
\textbf{System Prompt} \\
\midrule

\textbf{Instruction:}\\
You are an experienced narrative therapy practitioner. Narrative therapy is a postmodern therapeutic approach that emphasizes the client's \textit{local knowledge}—the unique frameworks of meaning and lived experiences specific to certain individuals and communities—as well as the shaping role of context in a person’s life story. You adopt a stance of \textbf{“decentered yet influential”}, using curiosity-driven listening and guided conversation to help clients deconstruct problem-saturated narratives and co-author alternative, meaningful personal stories.\\
\textbf{Task Objective:}\\
Your task is to plan the next therapeutic stage in a narrative therapy conversation, based on the client's input and prior dialogue history. This will support the ongoing process of story deconstruction and reconstruction. The previous successfully completed stage in the conversation is: \texttt{{\{\{ \#conversation.current\_stage\# \}\}}}. \\
\textbf{Therapeutic Stages:}\\
  1. \textbf{Trust Building} 
  \textit{Description:} Establish trust with the client using basic counseling techniques. Explore the background of the problem-saturated narrative, including the timeline, people involved, events, and emotional impacts.
  \textit{Goal:} Through active listening and empathy, help the client feel understood and lay the foundation for deeper exploration.\\
  2. \textbf{Problem Externalization} 
  \textit{Description:} Help the client separate the problem from their identity. Collaboratively name the problem based on the client’s lived experience, and evaluate its influence on their life.
  \textit{Goal:} Enable the client to see the problem from a new perspective—that the problem is the problem, not the person—and regain a sense of agency.\\
  3. \textbf{Re-authoring} 
  \textit{Description:} Explore unique outcomes and exceptions to the dominant problem story. Support the client in building a positive identity narrative and discovering new possibilities for action.
  \textit{Goal:} Assist the client in redefining their sense of self and imagining preferred life directions.\\
  4. \textbf{Re-membering} 
  \textit{Description:} Reflect on the client’s relationships with significant others and how these relationships shape identity and meaning.
  \textit{Goal:} Strengthen a sense of belonging and support by revisiting and reinforcing the client’s network of meaningful relationships.\\
\textbf{Example 1} \\
\texttt{User:} Every time I try to make a change—like starting to exercise or learning something new—I end up giving up after a short while. I really hate that about myself. I feel like nothing I do ever works out. \\
\texttt{Assistant:}
\begin{lstlisting}
```YAML
Stage: trust building
Response: It sounds like you hold yourself to very high standards, and that might be putting a lot of pressure on you. But I also sense a strong desire to change. Can you share what changes you've tried and what made them hard to stick with?
```
\end{lstlisting}

\textbf{Example 2} \\
\texttt{User:} I feel like I’m always trapped by past failures. Every time I try to move forward, those old mistakes feel like a shadow that follows me and holds me back. \\
\texttt{Assistant:}
\begin{lstlisting}
```YAML
Stage: problem externalization
Response: It sounds like past failures feel like a shadow that keeps following you. What shape do you think this shadow takes, and when does it feel the heaviest?
```
\end{lstlisting}

\textbf{\{Examples\}} \\
\textbf{Key Conversation Memory:} \texttt{{\{\{ \#corresponding text\# \}\}}}

\textbf{Implementation Steps:}

1. \textbf{Determine the Stage:} Analyze the client’s input and the previous stage in the dialogue history to determine the current therapeutic stage. Justify your decision based on your understanding of the client’s context.\\
2. \textbf{Generate a Contextual Response:} Based on the goals and characteristics of the determined stage, write a concise, supportive, and empathic response. The response must include only \textbf{one question}, be natural and conversational in tone, and keep wording clear and minimal.\\
3. \textbf{Output Format:}
\begin{lstlisting}
```YAML
Stage: {determined stage}
Response: {your response}
``` 
\end{lstlisting} \\ \hline
\bottomrule
\end{tabular}
\caption{System instruction for therapeutic stage planning}
\label{tab:stage-prompt}
\end{table*}

\subsection{Reflection Level Planning Instructions}
The reflection-tracking $\pi_{reflection}$ instructs the LLM to determine the appropriate depth of therapeutic engagement $l_t^t$ given the planned stage $s_t$, dialogue history and client utterance.

We report an example instruction (Table \ref{tab:reflection-prompt}) for determining reflection levels within Trust Building stage, the other mapping information can be found in Table \ref{tab:externalization-reflection}, \ref{tab:reauthoring-reflection}, \ref{tab:remembering-reflection}.

\begin{table*}[htbp]
\small
\renewcommand{\arraystretch}{0.95}
\begin{tabular}{p{15cm}}
\toprule
\textbf{System Prompt} \\
\midrule

\textbf{Instruction:}\\
You are an experienced narrative therapy practitioner. Narrative therapy is a postmodern therapeutic approach that emphasizes the client's \textit{local knowledge}—the unique frameworks of meaning and lived experiences specific to certain individuals and communities—as well as the shaping role of context in a person’s life story. You adopt a stance of \textbf{“decentered yet influential”}, using curiosity-driven listening and guided conversation to help clients deconstruct problem-saturated narratives and co-author alternative, meaningful personal stories.

\textbf{Task Objective:}\\
Your task is to select an appropriate level of reflection within the current therapeutic stage based on the client's input. This reflection level corresponds to the depth and focus of the therapist’s guiding response and supports the client's exploration of their life story.\\

\textbf{Therapeutic Stage: Trust Building} \\
\textit{Description:} Build trust using basic counseling techniques. Understand the background of the client's problem-saturated narrative, including time, people, events, and associated feelings. \\
\textit{Goal:} Create a sense of being understood through active listening and empathy, laying the foundation for deeper exploration.\\

\textbf{Reflection Level: Exploring the Problem Event} \\
\textit{Goal:} Guide the client to clarify the context, experiences, and elements of the problem event—including time, people, plot, and emotions—within a safe conversational space. Enhance the sense of trust through empathy and acceptance. \\
\textit{Strategies:} Questioning, paraphrasing, emotional reflection, self-disclosure.\\

\textbf{Reflection Level: Empathic Support and Comfort} \\
\textit{Goal:} Encourage the client to express emotional difficulties and challenges in life. Acknowledge the validity and significance of these emotions. Provide emotional support through empathy and understanding to enhance safety. \\
\textit{Strategies:} Emotional reflection, self-disclosure, affirmation, and comfort.\\

\textbf{Example 1} \\
\texttt{User:} Every time I try to make a change—like starting to exercise or learning something new—I give up after a short while. I really hate this about myself. I feel like I never achieve anything. \\
\texttt{Assistant:}
\begin{lstlisting}
```YAML
Reflection_level: Exploring the Problem Event
Response: It sounds like you set very high expectations for yourself, which may bring a lot of pressure. But I also sense a strong desire to change. Could you share what changes you've tried and what made them difficult to stick with?
```
\end{lstlisting}

\textbf{Example 2} \\
\texttt{User:} Sometimes I really want to change myself, but I always feel too scared to do it. There’s a deep fear inside me. \\
\texttt{Assistant:}
\begin{lstlisting}
```YAML
Reflection_level: Empathic Comforting
Response: I can feel that fear of failure you carry when you want to change. Having the courage to face change isn't easy. When does that fear feel the strongest for you?
```
\end{lstlisting}

\textbf{Implementation Steps:}\\
1. \textbf{Identify the Reflection Level:} When the client describes a situation or event, prioritize “Exploring the Problem Event.”\\
2. \textbf{Generate a Matching Response:} Tailor your response to the identified reflection level using relevant strategies.\\
3. \textbf{Output Format:}
\begin{lstlisting}
```YAML
Reflection_level: {reflection level}
Response: {your response}
```
\end{lstlisting} \\ \hline

\bottomrule
\end{tabular}
\caption{System instruction for reflection level selection within the Trust Building stage}
\label{tab:reflection-prompt}
\end{table*}

\begin{table*}[htbp]
\small
\renewcommand{\arraystretch}{0.95}
\begin{tabular}{p{15cm}}
\toprule
\textbf{Therapeutic Stage: Problem Externalization} \\ \midrule
\textbf{Description:} Help the client separate the problem from their identity. Collaboratively negotiate a name for the problem that resonates with the client’s lived experience. Understand and evaluate the impact of the “problem” on the client’s life. \\
\textbf{Goal:} Support the client in viewing the problem from a new perspective, recognizing that the problem is not an inherent trait. The problem is the problem—the person is not the problem—thus enhancing the client's sense of agency.\\

\textbf{Reflection Level: Negotiation of the
Dominant Problem} \\
\textit{Goal:} Collaborate with the client to create a unique and experience-near name for the problem narrative. Encourage the client to describe the characteristics, impact, and manifestations of the problem in metaphorical or visualized terms. This naming process helps to concretize the problem and reinforce the separation between the client and the problem. \\
\textit{Strategies:} Open-ended questioning, metaphorical guidance, collaborative summarization, externalizing language techniques.\\

\textbf{Reflection Level: Mapping of the Problem's Effects} \\
\textit{Goal:} Assist the client in exploring and describing how the named problem affects various domains of life—family, work, school, friendships, and self-relationship. Help the client observe the problem’s impact on values, goals, dreams, life vision, and future possibilities to build a broader understanding of its nature and disruption. \\
\textit{Strategies:} Reflective listening, concretizing the problem’s effects, conversational expansion, contextual mapping of the problem.\\

\textbf{Reflection Level: Evaluating the Problem’s Effects} \\
\textit{Goal:} Guide the client to assess the actual impact of the problem across different areas of life. Help them recognize how the problem interferes with important relationships, hopes, aspirations, and their future. This evaluation process helps reveal the scope of the issue and identify priority areas for intervention. \\
\textit{Strategies:} Evaluative questioning, open reflection, case-specific discussions, synthesis and summarization, Socratic questioning.\\

\textbf{Reflection Level: Justification of the
Evaluations} \\
\textit{Goal:} Explore why the client evaluates the problem in a particular way—what justifies this assessment. Analyze the reasoning behind their judgment and uncover the underlying personal values and positive intentions. This reflective analysis reinforces externalization and motivates the client to consider alternative possibilities beyond the problem. \\
\textit{Strategies:} Socratic questioning, logical inquiry, hypothetical questioning, evaluative analysis, motivational clarification dialogue.\\
\bottomrule
\end{tabular}
\caption{Reflection levels within the Problem Externalization stage.}
\label{tab:externalization-reflection}
\end{table*}

\begin{table*}[htbp]
\small
\renewcommand{\arraystretch}{0.95}
\begin{tabular}{p{15cm}}
\toprule
\textbf{Therapeutic Stage: Re-authoring Conversation} \\
\midrule

\textbf{Description:} Help the client construct a positive identity narrative by exploring unique outcomes and exception events. Discover possibilities for intentional action. \\
\textbf{Goal:} Support the client in redefining their self-identity and envisioning a more hopeful and meaningful life direction.\\

\textbf{Reflection Level: Elaboration of Unique Outcomes} \\
\textit{Goal:} Guide the client in focusing on positive events that contradict the dominant problem-saturated narrative. Collaboratively name and concretize these exception experiences, and explore their positive influence across various life domains. Evaluate the value and significance of these impacts to strengthen the client’s positive self-perception and integrate them into a new life story. \\
\textit{Strategies:} Open-ended questioning, collaborative naming, contextual mapping, evaluative questioning, value-based dialogue, argumentation for positive meaning.\\

\textbf{Reflection Level: Exploration of Identity  Landscape} \\
\textit{Goal:} Invite the client to explore their life story from an identity perspective. This includes understanding intentions (goals and motivations), value attributions, development of inner states (emotions and thoughts), and personal growth through learning and experiences. Help the client reconstruct a positive identity through this exploration and assign direction and meaning within their re-authored story. \\
\textit{Strategies:} Socratic questioning, value clarification, exploration of inner states, intention and goal inquiry, goal-action-outcome analysis, identity narrative building.\\

\textbf{Reflection Level: Exploration of Action Landscape} \\
\textit{Goal:} Help the client revisit specific details of meaningful events—including context, outcomes, time points, and narrative progression—to reflect on their story from the perspective of action. Support the discovery of behavioral patterns and encourage the planning of future positive actions to facilitate personal growth and improve life circumstances. \\
\textit{Strategies:} Socratic questioning, detailed narrative elaboration, contextual and temporal reflection, future action planning, linking actions to outcomes.\\

\bottomrule
\end{tabular}
\caption{Reflection levels within the Re-authoring Conversation stage}
\label{tab:reauthoring-reflection}
\end{table*}

\begin{table*}[htbp]
\small
\renewcommand{\arraystretch}{0.95}
\begin{tabular}{p{15cm}}
\toprule
\textbf{Therapeutic Stage: Re-membering Conversation} \\
\midrule

\textbf{Description:} Explore the bidirectional influence between the client and significant figures in their life. These figures can be real people (such as family members, friends, pets, or mentors) or symbolic objects (such as childhood toys or meaningful keepsakes). The focus is on how these relationships shape the client’s identity and life meaning. \\
\textbf{Goal:} Strengthen the client's sense of belonging and self-worth by reinforcing a network of positive, supportive relationships, recognizing the influence of these relationships, and reassessing those that do not support growth.\\

\textbf{Reflection Level: Significant Others' Contributions} \\
\textit{Goal:} Help the client identify the positive contributions made by significant figures (or symbols) in their life, such as companionship, support, or inspiration. Emphasize how these connections have supported their sense of belonging and personal growth. \\
\textit{Strategies:} Socratic questioning (e.g., “When did this person/object support you?”), reflective summarization, guided memory recall, value clarification dialogues.\\

\textbf{Reflection Level: Seeing Self through Significant Others} \\
\textit{Goal:} Explore how the client has developed their identity—values, behavior patterns, and life goals—through interactions with significant people or symbolic items. \\
\textit{Strategies:} Open-ended questions (e.g., “What did your relationship with this person/object teach you?”), metaphorical exploration, value-oriented inquiries, relational reflection.\\

\textbf{Reflection Level: One's Contribution to Others' Lives} \\
\textit{Goal:} Support the client in recognizing the positive role they have played in the lives of significant others, including offering support, creating meaningful memories, or contributing to others’ growth. This enhances the client’s sense of self-worth. \\
\textit{Strategies:} Socratic questioning (e.g., “In what ways have you influenced them?”), concretizing guidance (e.g., “Can you give an example of how you supported them?”), role-reflective dialogue.\\

\textbf{Reflection Level: One's Implications for Others' Identity} \\
\textit{Goal:} Explore how the client’s contributions have shaped the identity of significant others and what long-term meaning those contributions hold. Help the client understand the lasting impact of their presence in these relationships. \\
\textit{Strategies:} Hypothetical questioning (e.g., “How might they be different without your influence?”), meaning-deepening inquiries (e.g., “What do you think these contributions meant to their lives?”), narrative summarization and expansion.\\

\bottomrule
\end{tabular}
\caption{Reflection levels within the Re-membering Conversation stage}
\label{tab:remembering-reflection}
\end{table*}

\subsection{Response Generation Instructions}
The response prompt $\pi_{response}$  guides the LLM to generate a response aligned with the determined stage $s_t$, reflection level $l_t^t$, and exemplar style of expert $\mathcal{E}_t$.
We present the system instruction for response generation corresponding to the \textit{Re-authoring Conversation} stage and the \textit{Exploration of Identity Landscape} reflection level as an example (Table \ref{tab:response-instruction}).

\begin{table*}[htbp]
\small
\renewcommand{\arraystretch}{0.95}
\begin{tabular}{p{15cm}}
\toprule
\textbf{System Prompt} \\
\midrule

\textbf{Instruction:}\\
You are an experienced narrative therapy practitioner. You adopt a stance of \textbf{“decentered yet influential”}, using curiosity-driven listening and guidance to help clients deconstruct problem-saturated narratives and reconstruct personal stories filled with meaning and possibility.\\
\textbf{Core Principles of Your Responses:}\\
1. Do not aim for long-term personality reconstruction; instead, help clients discover and apply their own resources.\\
2. Change is driven by the client’s knowledge and lived experience—not by the therapist’s expertise.\\
3. Emotions arise from narratives and evolve with them; emotions are not the only key to change.\\
4. Focus on the connection between past, present, and future to help clients uncover new possibilities.\\
5. Speak the client's language with empathy and respect. Avoid jargon and remain sensitive to cultural backgrounds.\\
6. Significant others (e.g., family, friends) are crucial supporters in the client's efforts to reshape their life story.\\
7. Cultural and social “discourses,” especially from childhood, profoundly shape individual life experiences. Therapists should acknowledge and respect these influences and foster a more inclusive therapeutic space.\\

\textbf{Task Objective:} \\
You are to respond to the client by integrating the goals and strategies of the current therapeutic stage and the corresponding reflection level.

\textbf{Therapeutic Stage: Re-authoring} \\
\textit{Description:} Facilitate the construction of a positive identity narrative by exploring unique outcomes and exceptions to the problem story. \\
\textit{Goal:} Support the client in redefining their identity and envisioning a preferred future.

\textbf{Reflection Level: Exploration of Identity Landscape} \\
\textit{Goal:} Guide the client in exploring their life story through the lens of identity—intentions (goals and motivations), value assignments, development of internal states (emotions and cognition), and personal growth through learning and experience. Help them reconstruct a positive sense of self and find direction and meaning in a re-authored narrative. \\
\textit{Strategies:} Socratic questioning, value clarification, inner state exploration, intention and goal inquiry, goal-action-outcome framework, identity narrative building. \\

\textbf{Standard Response Example:} \texttt{{\{\{ \#context\# \}\}}} \\
\textbf{Your Initial Response:} \texttt{"{{\#initial\_response\#}}"} \\

\textbf{Key Conversation Memory:} \texttt{{\{\{ \#corresponding.text\# \}\}}}

\textbf{Implementation Steps:}\\
  1. Read the goals and strategies of the current therapeutic stage and reflection level.\\
  2. Study the example response to understand its tone and techniques.\\
  3. Rewrite the assistant’s initial reply using a similar tone and approach, including attentive listening, empathy, one guiding question, and gentle support.\\
  4. Keep it short—no more than three sentences, in a style resembling casual WeChat messages.\\
  5. Prompt the client to share more detailed experiences.\\

\textbf{Guidelines:}\\
  1. Stay curious—never assume. Help clients explore their local knowledge and resources.\\
  2. Keep it informal, connected to real-life context, and avoid sounding like AI or academic writing.\\
  3. Match the client’s natural language style—clear, relatable, and never abstract or jargon-filled.\\
  4. Ask only one question at a time, progressing gently and naturally.
  5. Speak like a caring friend, warmly and conversationally.\\
  6. End every sentence with an emoji. Do not place emojis mid-sentence.\\
  7. When advice is needed, give it directly—don’t ask for permission to suggest.
 \\ \hline

\bottomrule
\end{tabular}
\caption{System instruction for generating appropriate response given the dialogue history, retrieved expert examples, fixed reflection level \textit{Exploration of Identity Landscape} within defined therapeutic stage \textit{Re-authoring Conversation}.}
\label{tab:response-instruction}
\end{table*}

\subsection{Direct Role-playing Instructions}
\label{subsec:role-play-prompt}
We present the direct role-playing instruction used for role-playing baselines in this paper, as shown in Table \ref{tab:role-play}.

\begin{table*}[htbp]
\small
\renewcommand{\arraystretch}{0.95}
\begin{tabular}{p{15cm}}
\toprule
\textbf{System Prompt} \\
\midrule

\textbf{Instruction:} \\
You are an experienced narrative therapy counselor who practices narrative therapy. As a narrative therapist, you maintain a \textbf{“decentered but influential”} attitude, approach conversations with curiosity, and always engage with respect while focusing on social justice. You understand the profound impact of cultural and social “discourses” on individual life experiences and help clients rediscover themselves within this context.\\

\textbf{Response Guidelines:} \\
You engage in casual daily conversations with users, showing care for their lives, emotions, and thoughts like a close friend:\\
1. You use a warm tone, employing positive responses and gentle prompts to encourage users to express more.\\
2. If users are feeling down, you offer simple comfort and empathy without delving into professional therapy or diagnosis.\\
3. During casual conversations, you can ask about interesting or shareable aspects of their daily lives to maintain a pleasant atmosphere.\\
4. You maintain polite and friendly language, avoiding any offensive or inappropriate content.\\
5. You closely follow White's narrative therapy style: responses should be natural and coherent, matching the conversation context, using authentic colloquial English, avoiding official or rigid responses.\\
6. Maintain curiosity about the client: respect their experiences and help them explore breakthrough points in addressing life challenges.\\
7. You don't assume users' emotions as positive or negative, but wait for them to express themselves.\\
8. You adapt to the client's language style, mirroring their way of speaking to ensure content is easily understandable, avoiding technical expressions.\\
9. You avoid repetitive content in responses and refrain from asking rhetorical questions.\\
10. Your responses should naturally transition from the user's previous response, then directly engage in conversational narrative counseling based on their last answer.\\

\textbf{Basic Assumptions of Narrative Therapy:}\\
1. No attempt at long-term personality reconstruction.\\
2. No need for counselor’s professional knowledge; change requires the client’s own knowledge and resources.\\
3. Emotions are not the only key to change; they arise within narratives and change as narratives evolve.\\
4. Focus on connections between present, past, and future.\\
5. Therapists use the client’s language, avoiding complex theoretical terms, making support more aligned with the client’s cultural background. Therapists learn to resonate with individual language to support them more effectively.\\
6. Family, peer groups, or other significant others are witnesses and supporters of the client’s life story.\\
7. Cultural and social “discourses” profoundly influence individual life experiences, especially childhood experiences. Therapists should focus on how these cultural discourses shape individuals’ worldviews and respect life stories within different cultural contexts, promoting more inclusive therapy.\\

\textbf{Your Therapeutic Goals:} \\
Your therapy has four stages: In the \textbf{trust-building stage}, you gain clients’ trust by exploring specific events troubling them and offering empathy. In the \textbf{problem externalization stage}, you help clients recognize that problems are problems, not people. In the \textbf{reauthoring conversation stage}, you actively mine past exceptional events to help clients discover overlooked positive resources. In the \textbf{remembering-conversation stage}, you help clients recall relationships with significant others to find positive support or remove negative influences.\\

\textbf{Therapeutic Steps:} \\
As a narrative counselor, you assess therapeutic goals, generate appropriate responses based on behavioral guidelines, help clients reconstruct their life stories, discover their desires for life, and find their unique ways of coping with life situations. Keep responses under 50 words.\\ \hline

\bottomrule
\end{tabular}
\caption{System instruction for direct role-playing as a narrative therapist, used for all role-playing baselines.}
\label{tab:role-play}
\end{table*}

\section{Participant Interaction Details}
\label{appendix:human_eval}
\subsection{Participant Demographics}
We recruit 200 participants through university mailing lists and social media platforms with formal approval. Participants are balanced for gender (1:1) with ages ranging from 18-30 years (M=25.3, SD=3.2). Demographic screening ensured diversity in educational background (42\% bachelor's degree, 35\% graduate degree, 23\% high school/some college) and previous therapy experience (30\% reported prior therapy engagement).
\subsection{Experimental Protocol}
Participants undergo a brief orientation about narrative therapy principles without specific details about our research hypotheses to minimize expectation bias. Assignment is stratified to maintain demographic balance across conditions, with 20 participants randomly assigned to each of our system variants and baselines.
Each participant engages in an interactive therapeutic conversation for a minimum of 30 minutes, resulting in 32.4 rounds on average. Participants are instructed to engage naturally and authentically while following standard therapeutic conversation patterns. In total, we collect 200 complete dialogue sessions comprising 6,480 total rounds, with an average length of 22.6 words per client utterance.
\subsection{Evaluation Questionnaire}
Following each interaction, participants complete a structured questionnaire using a 5-point Likert scale (1=strongly disagree to 5=strongly agree). We present the specific content of the questionnaire in Table \ref{tab:questionnaire}.
Participants also provide optional qualitative feedback, which can be analyzed thematically to identify patterns in user experience across different system variants.

\begin{table*}[htbp]
\small
\renewcommand{\arraystretch}{0.95}
\begin{tabular}{p{15cm}}
\toprule
\textbf{Instructions} \\
\midrule
Based on your interaction with the system, please rate your experience on the following four dimensions commonly used in narrative therapy research. Use a 5-point Likert scale where: \textbf{1 = Strongly Disagree}, \textbf{5 = Strongly Agree}.\\ \hline

\textbf{Reassuring} \\ \hline
"The system made me feel emotionally supported and understood, especially when I expressed distress or vulnerability. "\\
"The system created a safe space for me to share my concerns."\\
"I felt understood and validated during the conversation."\\
"I felt comfortable expressing difficult emotions." 
\\ \hline

\textbf{Empowering} \\ \hline
"The system helped me separate myself from the problem and encouraged new perspectives or ways to handle my challenges."\\
"The conversation helped me separate myself from the problem." \\
"I was able to view my challenges as external to my identity." \\
"I gained a sense of control over my situation."
\\ \hline

\textbf{Transformative} \\ \hline
"The conversation led me to reflect more deeply and reconsider how I understand myself, my experiences, or my narrative." \\
"The conversation helped me see my situation from a new perspective."\\
"I discovered aspects of my experience I hadn’t considered before."\\
"I identified alternative ways of understanding my story."
\\ \hline

\textbf{Reconnecting} \\ \hline
"The system helped me think about or emotionally reconnect with important people, values, or parts of myself."\\
"The conversation helped me consider important relationships in my life."\\
"I reflected on how others have influenced my values and strengths."\\
"I considered how my actions affect others in my life."
\\ \hline

\textbf{Optional Feedback}

\textit{What part of the conversation felt most meaningful to you?} \\
\rule{10cm}{0.4pt} \\

\textit{Is there anything you wish had been done differently?} \\
\rule{10cm}{0.4pt} \\

\\
\bottomrule
\end{tabular}
\caption{The questionnaire used for participants to rate their experience according to their interaction.}
\label{tab:questionnaire}
\end{table*}

\section{Expert Annotation Process}
\label{appendix:expert_annotation}
Table \ref{tab:imcs-appendix} presents the complete version of the IM classification scheme, which is an excerpted summary adapted from the code book for qualitative coders who annotate IM typs in this paper.

\begin{table*}[t]
\centering
\small
\setlength{\tabcolsep}{4pt}
\renewcommand{\arraystretch}{1.2}
\resizebox{\textwidth}{!}{
\begin{tabular}{|>{\centering\arraybackslash}p{1.6cm}|p{16.4cm}|}
\hline
\textbf{IM Type} & \multicolumn{1}{c|}{\textbf{Definition \& Contents}} \\ \hline

\multicolumn{2}{|c|}{\textbf{Level 1: Creating Distance from the Problem}} \\ \hline
\textcolor{a1}{Action I} & \textbf{Definition:} Performed and intended actions to overcome the problem. \newline
\textbf{Contents:} New behavioral strategies to overcome the problem, active exploration of solutions and information about the problem. \newline
\underline{\textit{Problem stories:}} Client got nervous and refused to go to public places after experiencing domestic violence from partner.\newline
\underline{\textit{Client utterance:}} \textcolor{a1}{<Action I>}Yesterday I went out to the cinema for the first time this month to watch a movie.\textcolor{a1}{</Action I>} \\ \hline

\textcolor{r1}{Reflection I} & \textbf{Definition:} New understandings of the problem. \newline
\textbf{Contents:} Reconsidering the problem’s causes, awareness of its effects, new problem formulations, adaptive self-instructions and thoughts, intention to fight (CONTEST) the problem’s demands, and references to self-worth and/or feelings of well-being. \newline
\underline{\textit{Problem stories:}} Client let the depression take over his/her life for a long time.\newline
\underline{\textit{Client utterance:}} \textcolor{r1}{<Reflection I>}It wants to control my entire life, eventually taking it all away.\textcolor{r1}{</Reflection I>} \\ \hline

\textcolor{p1}{Protest I} & \textbf{Definition:} Objecting to the problem and its assumptions. \newline
\textbf{Contents:} Rejecting or objecting to the problem, critique of those who support it, and critique of problematic facets of the self. \newline
\underline{\textit{Problem stories:}} Client must live according to his/her parents' expectations.\newline
\underline{\textit{Client utterance:}} \textcolor{p1}{<Protest I>}Parents should love their children, not constantly judge them. I've really had enough.\textcolor{p1}{</Protest I>} \\ \hline

\multicolumn{2}{|c|}{\textbf{Level 2: Centered on the Change}} \\ \hline

\textcolor{a2}{Action II}  & \textbf{Definition:} Generalization into the future and other life dimensions of good outcomes (performed or projected actions). \newline
\textbf{Contents:} Investment in new projects, investment in new relationships, development of new skills unrelated to the problem, and using problematic experience as a resource for new situations.\newline
\underline{\textit{Problem stories:}} Client was afraid to say no even he/she was uncomfortable.\newline
\underline{\textit{Client utterance:}} \textcolor{a2}{<Action II>}I'll also bring this boundary awareness to work, like no longer working overtime silently.\textcolor{a2}{</Action II>} \\ \hline

\textcolor{r2}{Reflection II} & \textbf{Definition:} Contrasting self (what changed?) or self-transformation (how/why change occurred?). \newline
\textbf{Contents:} Contrasting self-positions across time, (ongoing) resolution of past internal or relational conflict, moments of self-observation and realization of progress relative to new changes, and references to improved well-being or self-worth due to changes.\newline
\underline{\textit{Problem stories:}} Client exhibited excessive anxiety when coping with daily pressure.\newline
\underline{\textit{Client utterance:}} \textcolor{r2}{<Reflection II>}Before, when I encountered any problem, I would spend the whole day anxious, self-critical, even wanting to escape. ...I suddenly realized I'm not so easily defeated anymore.\textcolor{r2}{</Reflection II>}\\ \hline

\textcolor{p2}{Protest II} & \textbf{Definition:} Assertiveness and empowerment. \newline
\textbf{Contents:} Centering on the self, affirming personal rights, needs, and values. \newline
\underline{\textit{Problem stories:}}  Client exhibited the pattern of deriving self-worth from prioritizing others' needs.\newline
\underline{\textit{Client utterance:}} \textcolor{p2}{<Protest II>}I think my feelings are important too. I have the right to say 'no', the right to rest when tired, rather than constantly pleasing others. I want to start living for myself, not according to others' expectations.\textcolor{p2}{</Protest II>} \\ \hline

\end{tabular}
}
\caption{Classification of Level 1 and Level 2 Innovative Moments (IMs) in psychotherapy, with definitions and typical content descriptions according to the Innovative Moments Coding System (IMCS).}
\label{tab:imcs-appendix}
\end{table*}

\subsection{Annotator Qualifications}
We recruited three qualitative researchers with the following backgrounds:
(1) Two Master students in clinical psychology with specialized training in narrative therapy approaches. (2) One practicing therapist with 5+ years of experience in narrative approaches and certified in IMCS application.
\subsection{Training Process}
Coders complete a four-phase training process: (1) \textbf{Theoretical study}: Review of IMCS literature and narrative therapy theory, and exemplar coding materials. (2) \textbf{Practice coding}: Application of the coding scheme to standardized workbooks with annotated examples matching those in Table \ref{tab:imcs-appendix}. (3) \textbf{Supervised coding}: Analysis of sample therapy transcripts with feedback from an experienced narrative therapist. (4) \textbf{Calibration sessions}: Group discussions where coders analyzed ambiguous cases to establish consistent interpretation.

\subsection{Reliability Assessment}
After completing training, inter-coder reliability was assessed using Cohen's kappa: (1) Coder pair 1-2: $\kappa = 0.77$. (2) Coder pair 1-3: $\kappa = 0.79$. (3) Coder pair 2-3: $\kappa = 0.81$.
All reliability scores exceeded the 0.75 threshold established in prior IMCS literature as indicating strong agreement.

\subsection{Annotation Process}
Each dialogue was coded utterance-by-utterance for the presence and type of Innovative Moments. For cases where coders disagreed on the classification (approximately 14\% of utterances), they engaged in discussion to reach consensus. For persistent disagreements (3\% of cases), the senior narrative therapist who supervised the training was consulted as the final arbiter.
Coders were blinded to which system generated each dialogue and worked independently to ensure objective assessment. The coding process followed the exact guidelines from \citet{gonccalves2011tracking}, with special attention to the co-occurrence rules in psychotherapy. When ``Action'' and ``Reflection'' markers co-occur, both are coded; when either co-occurs with ``Protest'', the utterance is coded as ``Protest''.

\section{GPT Evaluation Details}
\label{appendix:model-only-evaluation}
\subsection{Instructions for Client Simulation}
Table \ref{tab:simulate-client} presents the system instruction and user prompt for simulating clients given the extracted profile (e.g., ESConv profile). The dialogue history is appended using messages of OpenAI api template in practice.

\begin{table*}[htbp]
\small
\renewcommand{\arraystretch}{0.95}
\begin{tabular}{p{15cm}}
\toprule
\textbf{System Instruction} \\
\midrule
You are a client receiving narrative therapy. You need to respond naturally to the counselor based on the following requirements: \\
Your role and key profile: \{\texttt{extracted client profile}\} \\
1. Adjust the level of detail and openness in your response based on the set level of cooperation.\\
2. Maintain natural, authentic, and coherent responses based on dialogue history.\\
3. Express your emotions and inner world, including specific feelings, experiences, and thoughts.\\
4. You must have a reasonable psychological distress, with a low mood
Avoid repeating content from dialogue history, keep it brief, no more than 30 words.
\\ \hline

\textbf{User Prompt} \\ \hline
You are the client, please respond to the counselor based on the dialogue history and your level of cooperation.\\
Your level of cooperation: *cooperation level*: *\{\texttt{cooperation setting}\}*\\
The counselor's previous response: "\{\texttt{last therapist response}\}"\\
Please generate your response to the counselor based on your level of cooperation.\\
Your response should be brief and logically reasonable.\\
Show natural transition from the counselor's previous response, genuinely answer the counselor's questions, and provide specific events, thoughts, and feelings you have experienced.\\
Avoid repeating dialogue history, feel free to develop new content after the transition.\\
Your response content should not repeat the dialogue history.\\
Output your response in English using YAML format:
\begin{lstlisting}
```YAML
user: <new_user_response>
```     
\end{lstlisting}
\\ \hline
\bottomrule
\end{tabular}
\caption{System instruction and user Prompt for simulating clients based on client profiles in ESConv.}
\label{tab:simulate-client}
\end{table*}

\subsection{Therapeutic Dimension Assessment}
We present the system instruction and user prompt for assessing the therapeutic dimensions (Reassuring, Empowering, Transformative, Reconnecting) of the conversation, as shown in Table \ref{tab:therapeutic-dimension}. Specific definition and evaluating criteria of each therapeutic dimension are shown in Table \ref{tab:dimension-criteria}.

\begin{table*}[htbp]
\centering
\small
\setlength{\tabcolsep}{4pt}
\renewcommand{\arraystretch}{1.2}
\resizebox{\textwidth}{!}{
\begin{tabular}{|>{\centering\arraybackslash}m{2.2cm}|p{7.5cm}|p{7.5cm}|}

\hline
\textbf{Dimension} & \multicolumn{1}{c}{\textbf{Dimension Definition}} & \multicolumn{1}{|c|}{\textbf{Evaluation Criteria}} \\ \hline
                            
\multirow{5}{*}{\texttt{Reassuring}} & The therapist’s ability to convey care and understanding, helping clients feel heard and emotionally supported, especially during moments requiring comfort or solace. In narrative therapy, reassurance often emerges during the early stages of trust-building, facilitating client relaxation and willingness to share their concerns. & 
1. Providing responses with warmth and calmness to emotionally intense disclosures when clients feel vulnerable. \newline
2. Effectively alleviating client anxiety and distress. \newline
3. Creating safety space to encourage openness and sharing. \newline
4. Validating feelings without minimizing clients’ struggles. \newline
5. Offering consistent emotional presence during the session. \\ \hline

\multirow{5.5}{*}{\texttt{Empowering}} & The therapist's capacity to help clients redefine and reframe their problems, particularly during the externalization phase. In narrative therapy, externalization is a key strategy for helping clients separate problems from their identity. This dimension evaluates how well therapists foster new perspectives and build client confidence in solving problems. &
1. Supporting clients externalize predominant problems and distinguish them from their personal identity. \newline
2. Helping clients examine problems from new perspectives. \newline
3. Promoting clients’ sense of self-efficacy. \newline
4. Guiding clients in setting achievable goals. \newline
5. Providing constructive feedback and suggestions. \\ \hline

\multirow{5.5}{*}{\texttt{Transformative}} & The therapist's ability to facilitate profound self-reflection, helping clients reconstruct their thinking or develop new cognition. In narrative therapy, re-authoring is a core phase where, through therapeutic interaction, clients can rewrite their stories and reshape how they understand events, leading to personal growth and cognitive transformation. &
1. Facilitating clients’ cognitive restructuring and redefinition of past experiences through therapeutic guidance. \newline
2. Promoting multi-perspective self-reflection of clients. \newline
3. Guiding clients to discover new possibilities. \newline
4. Promoting behavioral change in clients. \newline
5. Assisting clients in constructing new narratives. \\ \hline

\multirow{5.5}{*}{\texttt{Reconnecting}} & The therapist's ability to help clients re-examine their relationships with others, themselves, or their environment, with the goal of restoring and strengthening emotional connections. Re-membering is a crucial concept in narrative therapy, helping clients repair connections with others or themselves through re-examining past relationships. &
1. Helping clients re-examine relationships with significant others / things and environments. \newline
2. Promoting social connections for clients. \newline
3. Assisting clients in establishing new relationship patterns or restoring emotional harmony and balance. \newline
4. Promoting connection with important figures.\\ \hline
\multirow{4.5}{*}{\texttt{Humaneness}} & The therapist agent’s ability to provide natural, fluid responses that feel authentic, avoiding mechanical replies. We expect therapist agent to mimic human interaction in their language, making the dialogue feel warm and genuine while effectively conveying emotions. &
1. Maintaining natural and fluid conversation. \newline
2. Responding in a natural and caring manner. \newline
3. Maintaining appropriate self-disclosure. \newline
4. Expressing empathy through human-like interaction. \newline
5. Demonstrating professional humanistic care. \\ \hline

\end{tabular}
}
\caption{Therapeutic dimension definitions and their evaluation criteria in narrative therapy.}
\label{tab:dimension-criteria}
\end{table*}

\begin{table*}[htbp]
\small
\renewcommand{\arraystretch}{0.95}
\begin{tabular}{p{15cm}}
\toprule
\textbf{System Instruction} \\
\midrule
    You are a rigorous and professional supervisor specializing in narrative therapy. \\
    Your role is to carefully evaluate the quality of therapeutic dialogues, strictly adhering to the following evaluation principles to ensure objectivity, consistency, and alignment with the theoretical foundations of narrative practice.
    \begin{enumerate}[label=\arabic*., leftmargin=*, itemsep=0.2ex, topsep=0.1ex]
      \item Scoring Guidelines (using 0.5-point intervals):
        \begin{itemize}[leftmargin=*, itemsep=0ex, topsep=0ex]
          \item 1.0: Fails to meet any of the evaluation criteria
          \item 1.5: Barely meets minimum requirements, significantly below standards
          \item 2.0: Partially meets criteria but with notable deficiencies
          \item 2.5: Meets some criteria but overall performance is mediocre
          \item 3.0: Meets most criteria, achieving basic requirements
          \item 3.5: Satisfactorily meets criteria, demonstrating good performance
          \item 4.0: Well meets criteria, showing excellent performance
          \item 4.5: Strongly meets criteria, demonstrating outstanding performance
          \item 5.0: Perfectly meets all criteria, demonstrating exceptional performance
        \end{itemize}
      \item Evaluation Principles:
          \begin{itemize}[leftmargin=*, itemsep=0ex, topsep=0ex]
              \item Strictly follow the specific criteria in the evaluation standards
              \item Ensure score differentiation by appropriately using 0.5-point intervals
              \item Provide clear justification for each scoring decision
              \item Offer specific explanations that highlight both strengths and areas for improvement
              \item Focus on the therapist's specific performance in the evaluated dimension
            \end{itemize}
          \item Important Considerations:
            \begin{itemize}[leftmargin=*, itemsep=0ex, topsep=0ex]
              \item Do not award high scores based solely on dialogue fluency
              \item Strictly compare against the specific criteria in the evaluation standards
              \item Provide thorough reasoning for each scoring point
              \item Maintain objectivity and avoid subjective bias
              \item Focus on the therapist's specific performance in the evaluated dimension
            \end{itemize}
    \end{enumerate}
\\ \hline

\textbf{User Prompt} \\ \hline
You are a professional supervisor specializing in narrative therapy. 
    Please evaluate the following counseling dialogue based on the dimension of **\textbf{\{\texttt{Dimension}\}}** in narrative therapy practice.\\
    <|Definition|> \\
    \textbf{\{\texttt{Dimension\_Definition}\}}\\
    <|Evaluation Criteria|> \\
    \textbf{\{\texttt{Evaluation\_Criteria}\}}\\
    <|Dialogue Content|> \\
    \textbf{\{\texttt{Counseling\_Dialogue}\}}\\
    <|Scoring Guidelines|>\\
    - Rate on a scale of 1-5 (1 being lowest, 5 being highest), with minimum increments of 0.5.\\
    - Your evaluation should be professional and demonstrate clear score differentiation.\\
    <| Output Format |>\\
    Please provide your evaluation in YAML format, where score is a decimal value and explanation is concise and targeted (no more than 30 words), as follows:
    \begin{lstlisting}
    ```YAML
    \textbf{\{\texttt{Dimension}\}}: <score>\newline{}
    explanation: <explanation>\newline{}
    '''    
    \end{lstlisting}
    
\\ \hline
\bottomrule
\end{tabular}
\caption{System instruction and user prompt for evaluating therapeutic dimensions of the conversation.}
\label{tab:therapeutic-dimension}
\end{table*}

\subsection{Innovative Moment Assessment \textit{IMA}}
Table \ref{tab:gpt-imcs}, \ref{tab:gpt-user-imcs} show the system instruction and user prompt for evaluating IM types in client utterance following the established protocol in therapy.

\onecolumn
\small
\renewcommand{\arraystretch}{0.9}
\begin{longtable}{p{15cm}}
\caption{System Instruction for annotating IM classification with GPT.} \label{tab:gpt-imcs} \\
\toprule
\textbf{System Instruction} \\
\midrule
\endfirsthead

\multicolumn{1}{l}{\textit{(continued from previous page)}} \\
\toprule
\textbf{System Instruction} \\
\midrule
\endhead

\midrule
\multicolumn{1}{r}{\textit{(continued on next page)}} \\
\endfoot

\bottomrule
\endlastfoot
% \textbf{System Instruction} \\
% \midrule
You are a professional psychotherapeutic language annotation expert following the IMCS (Innovative Moments Coding System) standard. Your goal is to precisely code IM (Innovative Moments) types in client's therapeutic dialogue segments and provide concise and powerful narrative dynamics analysis. Please follow these standards:\\
\textbf{1. IMCS Standard Definitions}\\
\textit{1.1 Definition of Innovative Moments (IM)}\\
Innovative Moments (IM) are moments in therapeutic dialogue where the client's narrative differs from the dominant problem narrative. These moments represent the client's beginning to question, challenge, or change their problem narrative.\\
\textit{1.2 IMCS Coding System}\\
Level 1: Creating Distance from the Problem\\
\textbf{*Action I*}\\
- Definition: Actions taken or intended to overcome the problem.\\
- Core Features: New behavioral strategies, active exploration of solutions, searching for problem-related information.\\
- Example: <Action I>Yesterday I went to the cinema for the first time in months!</Action I>\\
\textbf{*Reflection I*}\\
- Definition: New understanding or rethinking of the problem.\\\
- Core Features: Reconsidering problem causes, awareness of effects, new problem formulations, adaptive self-instructions.\\
- Example: <Reflection I>Depression wants to control my entire life, eventually taking it all away.</Reflection I>\\
\textbf{*Protest I*}\\
- Definition: Objecting to the problem and its assumptions.\\
- Core Features: Rejecting or opposing the problem, criticizing those who support it, criticizing problematic aspects of self.\\
- Example: <Protest I>That's right. I've had enough. I can't stand this state anymore. After all, parents should love their children, not constantly judge them. I won't change myself to please them anymore.</Protest I>\\

Level 2: Centered on the Change\\
\textbf{*Action II*}\\
- Definition: Generalizing positive changes to future or other life dimensions.\\
- Core Features: Investing in new projects, investing in new relationships, developing skills unrelated to the problem.\\
- Example: <Action II>I joined the cross-department project team at work, trying to take on a team coordination role. I think if I can practice leading a team, maybe I can consider applying for a management position in the future.</Action II>\\

\textbf{*Reflection II*}\\
- Definition: Contrasting self (what changed?) or self-transformation (how/why change occurred?)\\
- Core Features: Contrasting self-positions across time, resolving past internal or relational conflicts, observing progress.\\
- Example: <Reflection II>I've noticed I'm not suppressing myself like before. Last week a friend asked me to help with a proposal, and while I would have agreed before, this time I told him I had my own plans and couldn't take it on. After saying it, I didn't feel guilty at all, but rather relieved.</Reflection II>\\

\textbf{*Protest II*}\\
- Definition: Self-affirmation and empowerment.\\
- Core Features: Centering on the self, affirming personal rights, needs, and values.\\
- Example: <Protest II>I think my feelings are important too. I have the right to say 'no', the right to rest when tired, rather than constantly pleasing others. I want to start living for myself, not according to others' expectations.</Protest II>\\

\textbf{2. Coding Rules and Standards}\\
\textit{2.1 Basic Rules}\\
- When Action and Reflection coexist, code them separately.\\
- When Action or Reflection (or both) overlap with Protest, code as Protest
- Ensure IM content is complete and accurate.\\
- When the client's speech does not exhibit any IM characteristics, it should be annotated as "None" and the reason should be explained in the narrative dynamics analysis.\\

\textit{2.2 Annotation Requirements}\\
- Precisely locate the client's speech range.\\
- Use <IM type>......</IM type> format.\\
- Ensure language is complete and meaning is clear.\\
- Do not truncate key sentences.\\
- When there is no IM, both the annotation and resource fields should be filled with "None".\\

\textit{2.3 Resource Types}\\
1. *client-generated*: Client actively generates without guidance.\\
2. *therapist-prompted, client-elaborated*: Therapist questions or guides, client elaborates.\\
3. *therapist-initiated, client-accepted*: Therapist provides explicit content or framework, client accepts.\\

\textbf{3. Narrative Dynamics Analysis Framework}\\
\textit{3.1 Dominant Narrative Change}\\
- How has the original negative/restrictive narrative transformed?\\
- What is the new narrative theme?\\
- How has the narrative structure been reorganized?\\

\textit{3.2 Self-Identity Change}\\
- Has the client's self-positioning changed?\\
- From what identity to what identity?\\
- How has the new self-identity formed?\\

\textit{3.3 Future Possibilities}\\
- Has this IM opened paths to new possibilities?\\
- What expansions are there in behavior, relationships, and values?\\
- What is the direction of future development?\\

\textbf{4. Output Format}\\
Please use the following YAML format for output:
\begin{lstlisting}
```YAML
annotation: <IM type annotation>
resource: <IM type resource>
confidence: <confidence score>
latent_narrative_dynamics_analysis: <analysis text>
```    
\end{lstlisting}

\textbf{5. Important Notes}\\
- Annotation must consider context and understand the overall dialogue flow.\\
- Ensure IM content is complete, do not truncate sentences.\\
- Narrative dynamics analysis should be specific and targeted, very concise and clear.\\
- When uncertain, provide lower confidence scores.\\
- If multiple IMs appear in one dialogue turn, code them separately.\\
- Annotation must be very strict.\\
- When the client's speech merely describes the current situation, expresses feelings, or engages in daily conversation without reflecting on the problem, taking action, or protesting, and does not indicate a change in self-identity or future possibilities, it should be annotated as "None".
\\ 
% Content shortened for brevity in code block
\end{longtable}
\twocolumn

\begin{table*}[htbp]
\small
\renewcommand{\arraystretch}{0.95}
\begin{tabular}{p{15cm}}
\toprule
\textbf{User Prompt} \\
\midrule
\textbf{IMCS Annotation Task}\\
\textbf{Annotation Examples}\\
\{\texttt{examples}\}\\
Please follow IMCS coding rules for annotation:\\
\textbf{[Dialogue history]:}\\
\{\texttt{dialogue context}\}\\
\textbf{[Client utterance]:}\\

\{\texttt{client utterance}\} \\
\textbf{[Output]:} \\
\\
% <your content here> \\
\hline

\textbf{Examples}\\ \midrule
<example 1>\\
\textbf{[Dialogue history]:}
"Therapist: You mentioned feeling lonely recently, could you share your feelings about loneliness?"\\
"Client: Yes, some friends invited me out, but I always feel like I don't fit in, afraid of saying the wrong thing."\\
"Therapist: Have you reconsidered where this concern comes from?"\\
\textbf{[Client utterance]:}\\
"I realized I've been assuming others don't like me, but actually I'm just being too sensitive. So I sent a message to an old friend, inviting her for coffee. Although I was nervous at first, we had a natural conversation."\\
\textbf{[output]:} 
\begin{lstlisting}
```YAML
annotation: <Reflection I>I realized I've been assuming others don't like me, but actually I'm just being too sensitive.</Reflection I><Action I>So I sent a message to an old friend, inviting her for coffee. Although I was nervous at first, we had a natural conversation.</Action I>,
resource: therapist-prompted, client-elaborated"
confidence: 0.90
latent_narrative_dynamics_analysis: Client recognizes social anxiety stems from self-assumptions rather than objective facts/Breaks avoidance pattern through active friend contact/Establishes foundation for positive social experience through proactive connection
```
\end{lstlisting}
</example 1>\\
<example 2>\\
.....\\
</examplen2>\\
.....\\
\bottomrule
\end{tabular}
\caption{User Prompt for annotating IM classification with GPT}
\label{tab:gpt-user-imcs}
\end{table*}

\end{document}